\documentclass[journal]{IEEEtran}
\pdfoutput=1

\usepackage{ifpdf}
 \ifpdf
 \else
 \fi
\usepackage{animate}
\usepackage{capt-of}% or \usepackage{caption}

\ifCLASSOPTIONcompsoc
  \usepackage[nocompress]{cite}
\else
  \usepackage{cite}
\fi
\ifCLASSINFOpdf
   \usepackage[pdftex]{graphicx}
   \graphicspath{{./figArXiv/}}
   \DeclareGraphicsExtensions{.pdf,.jpeg,.jpg,.png}
\else
\fi
\usepackage{color}

\usepackage{subfigure}
\usepackage{hyperref}

\begin{document}

\title{Computational Model for Predicting Visual Fixations from Childhood to Adulthood}
%\title{Visual attention saccadic models learn to emulate gaze patterns from childhood to adulthood}

%~\IEEEmembership{Life~Fellow,~IEEE}% <-this % stops a space

\author{Olivier~Le~Meur,
        Antoine~Coutrot,
        Zhi~Liu,
        Adrien~Le~Roch,        
        Andrea~Helo,
        Pia~Rama        
        % <-this % stops a space
\IEEEcompsocitemizethanks{\IEEEcompsocthanksitem M. Le Meur and M. Le Roch are with University of Rennes 1 IRISA, France. E-mail: olemeur@irisa.fr\protect\\
\IEEEcompsocthanksitem Ms. Helo and Ms. Rama are with University of Paris Descartes. 
\IEEEcompsocthanksitem M. Liu is with Shanghai University. 
\IEEEcompsocthanksitem M. Coutrot is with University College London.}% <-this % stops an unwanted space
}


\markboth{Report 2017: Computational Model for Predicting Visual Fixations from Childhood to Adulthood}%
{Le Meur \MakeLowercase{\textit{et al.}}}



\IEEEtitleabstractindextext{%
\begin{abstract}
How people look at visual information reveals fundamental information about themselves, their interests and their state of mind. While previous visual attention models output static 2-dimensional saliency maps, saccadic models aim to predict not only where observers look at but also how they move their eyes to explore the scene. Here we demonstrate that saccadic models are a flexible framework that can be tailored to emulate observer's viewing tendencies. More specifically, we use the eye data from 101 observers split in 5 age groups (adults, 8-10 y.o., 6-8 y.o., 4-6 y.o. and 2 y.o.) to train our saccadic model for different stages of the development of the human visual system. We show that the joint distribution of saccade amplitude and orientation is a visual signature specific to each age group, and can be used to generate age-dependent scanpaths. Our age-dependent saccadic model not only outputs human-like, age-specific visual scanpath, but also significantly outperforms other state-of-the-art saliency models. In this paper, we demonstrate that the computational modelling of visual attention, through the use of saccadic model, can be efficiently adapted to emulate the gaze behavior of a specific group of observers.
\end{abstract}

}

\hypersetup{
pdftitle={\@title},
pdfauthor={\@author},
colorlinks=true, %colorise les liens
breaklinks=true, %permet le retour à la ligne dans les liens trop longs
urlcolor= blue, %couleur des hyperliens
linkcolor= blue, %couleur des liens internes
citecolor=blue,    %couleur des liens de citations
bookmarksopen=true,
pdftoolbar=false,
pdfmenubar=false,
}

\maketitle

\IEEEdisplaynontitleabstractindextext

% \fi
\IEEEpeerreviewmaketitle

\section{Introduction}\label{sec:introduction}

\IEEEPARstart{O}{culus} \textit{animi index} is an old Latin proverb that could be translated as \textit{the eyes reflect our thoughts}. Eye-movements, revealing how and where observers look within a scene, are mainly composed of fixations and saccades. Fixations aim to bring areas of interest onto the fovea where the visual acuity is maximum. Saccades are ballistic changes in eye position, allowing to jump from one position to another. Visual information extraction essentially takes place during the fixation period. The sequence of fixations and saccades an observer performs to sample the visual environment is called a visual scanpath.

Thanks to the advent of modern eye-trackers, allowing us to capture gaze with a high spatial and temporal resolution, a large amount of eye tracking data can be collected with a relative simplicity. Given that the execution of eye movements is the result of a complex interaction between various cognitive processes, mining eye tracking data may provide many indications on our personality, on our mood, and more generally speaking, on the cognitive states of our mind.  The way we explore our environment, the way we moves our eyes from one location to another in order to inspect it accurately, may reveal information about our cognitive state. For instance, Henderson et al.~\cite{Henderson2013} inferred the task the participants are engaged in by analyzing eye-movements. Wang et al.~\cite{Wang2015} combined eye tracking with computational attention models  in order to screen for mental diseases such as autism spectrum disorder (see also~\cite{Tseng2013,Itti2015}). Tavakoli et al.\cite{Tavakoli2015} investigated the influence of eye-movement-based features to determine the valence of images. 
%As emphasized by~\cite{Tseng2013}, recording eye movements of observers while free-viewing stimuli onscreen has many advantages, such as the relative simplicity and the ability to collect a large amount of data (see~\cite{Itti2015} for more details).

\begin{figure*}[!t]
\centering
\includegraphics[width=\linewidth]{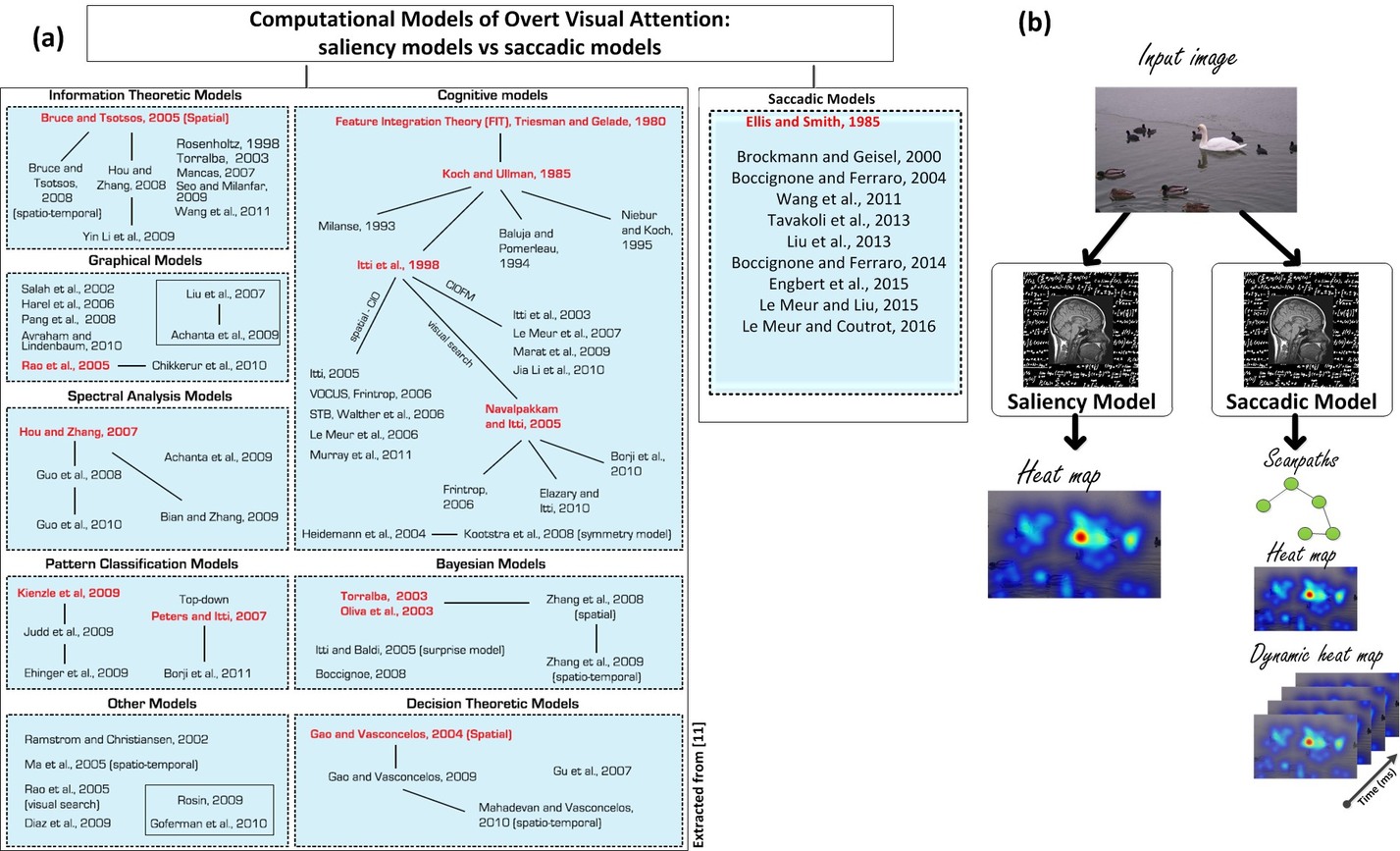}
\caption{(a) taxonomy of computational models of overt visual attention. Compared to previous taxonomy proposed by~\cite{Borji2013}, a second category featuring saccadic models is added. (b) The conceptual difference between saliency models and saccadic models is illustrated. Saliency models output a 2D static saliency map (or heat map) whereas saccadic models compute visual scanpaths from which static as well as dynamic saliency maps can be computed.}
\label{fig:classif}
\end{figure*}

Predicting where we look at within a scene is of particular relevance for many computer vision applications such as computer graphics~\cite{Longhurst2006gpu}, quality assessment~\cite{Ninassi2007,Liu2011} and compression~\cite{Li2011} to name a few. There exist many computational models of overt visual attention. In 2013, Borji and Itti~\cite{Borji2013} have proposed a taxonomy of saliency models. We propose to update this classification by making the distinction between saliency model and saccadic model, as illustrated in~\figurename~\ref{fig:classif}.

Saliency models aim to predict the salience of a visual scene. They are based on low-level visual features including color, intensity, and orientation. They process these visual features at several scales using center-surround differences. This process filters out redundant information and outputs feature maps, one per channel. A final saliency map is obtained by combining these feature maps. In contrast with saliency models, saccadic models intend to predict the sequence of eye fixations, i.e. the fashion an observer deploys his/her gaze while viewing a stimulus on screen. Rather than computing an unique saliency map, saccadic models compute visual scanpaths from which scanpath-based saliency maps can be computed.  This is illustrated in~\figurename~\ref{fig:classif} (b). As discussed later in this paper, saccadic models offer many advantages over saliency models. The most important one is the ability to tailor the saccadic model to a particular context, such as a particular type of scene, a particular population or to a particular task at hand~\cite{LeMeurSPIE2016}.

Modelling the human visual attention is a complex task, because of the number of underlying biological mechanisms involved in the visual perception. One of the major difficulties is the high variability in eye-movements. This dispersion is due to many factors, which could be related for instance to the task at hand~\cite{Yarbus1967}, the cultural heritage~\cite{Nisbett2003}, the gender~\cite{Miyahira2000,Coutrot2016} and observers' age~\cite{dowiasch2015effects}. The last factor, i.e. the age of observer, is the central concern of this paper. 

In this paper, we aim to design an age-dependent saccadic model in order to reproduce the gaze behavior of a certain target age group. This study is based on a reliable scanpath signature shared within each age group. This signature is then used to emulate the gaze behavior of a specific age group of observers.

This report is organized as follows. Section~\ref{sec:saccadicModel} presents saccadic models and focuses more specifically on the modelling framework proposed in~\cite{LeMeur2015,LeMeur2016}. We will also stress how we can tailor this saccadic model for different age groups. Section~\ref{sec:Helo} presents the eye tracking dataset that is used to determine the scanpath-based signature~\cite{Helo2014}. Section~\ref{sec:adaptationSaccadicModel} presents the age-dependent saccadic model and section~\ref{sec:perf} evaluates its performances. In Section~\ref{sec:conclusion}, we discuss the results and draw some conclusions.

%****************************************************
\section{Saccadic model}\label{sec:saccadicModel}
%****************************************************
\subsection{Definition}
%In this project, which is still in its incipient stage
Saccadic models aim to generate plausible visual scanpaths, i.e. the actual sequence of fixations and saccades an observer would do while viewing stimuli onscreen. By the term plausible, we mean that the predicted scanpaths should be as similar as possible to human scanpaths. They should exhibit similar characteristics, such as the same distributions of saccade amplitudes and saccade orientations. In summary, a saccadic model must predict how  observer moves his gaze, but also where the observer looks. 

% , fixation points should fall within the most salient areas. 

Most existing saccadic models assume that gaze shifts follow a Markov process, meaning that the next gaze location depends only on the current one. In 2000, Brockmann and Geisel~\cite{Brockmann2000} generated a sequence of fixation points by considering a stochastic jump process, in which the transition probability density of shifting the gaze from one fixation to another depends on the product of a random salience field and the amplitude of the generated saccade. Boccignone and Ferraro~\cite{Boccignone2004} extended Brockmann's work, and modeled eye gaze shifts by using L{\'e}vy flights constrained by a bottom-up saliency map. Wang et al.~\cite{Wang2011} used the principle of information maximization to generate scanpaths on natural images. One interesting point is that they learned the distribution of saccade amplitudes from their own eye movements dataset in order to constrain the selection of the next fixation point. Liu et al.~\cite{Liu2013semantically} went further by using a Hidden Markov Model (HMM) with a Bag-of-Visual-Words descriptor of image regions to account for semantic content.
Tavakoli et al.~\cite{Tavakoli2013} also incorporates visual working memory as well as a Gaussian mixture to estimate the  distribution of saccade amplitudes. Engbert et al.~\cite{Engbert2015spatial} propose the \textit{SceneWalk} model of scanpath generation based on two independent processing streams for excitatory and inhibitory pathways. Both are represented by topographic maps: the former represents the foveated saliency map  whereas the latter is used for inhibitory tagging. These two maps for attention and inhibitory tagging are then combined. The next fixation point is selected thanks to a stochastic selection~\cite{Engbert2015spatial}. In~\cite{LeMeur2015,LeMeur2016}, Le Meur et al. propose a model of scanpath generation by considering spatially-variant and context-dependent joint distribution of saccade amplitudes and orientations. The next subsection underlines its main components.

\begin{figure}[!t]
\centering
\includegraphics[width=\linewidth]{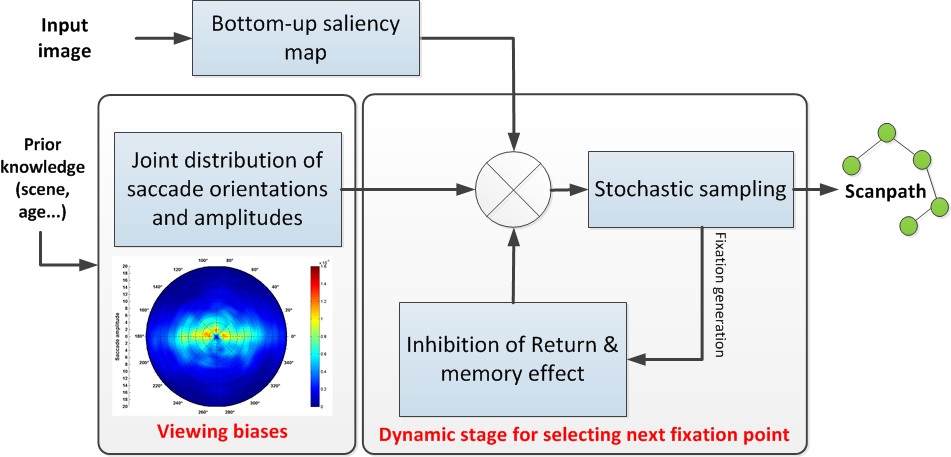}
\caption{Flow chart of the saccadic model proposed in~\cite{LeMeur2015,LeMeur2016}. The model takes as input: the original image as well as prior information related to the type of the scene, the age of observers, etc. It outputs a set of visual scanpaths.}
\label{fig:schema}
\end{figure} 

\subsection{Le Meur's saccadic model}
Predicted scanpaths result from the combination of three components, namely a bottom-up saliency map, viewing biases and memory mechanism, as illustrated by~\figurename~\ref{fig:schema}. In the following, we summarize the main operations involved in the method proposed in~\cite{LeMeur2015,LeMeur2016}.

Let $\mathcal{I}:\Omega\subset\mathcal{R}^2\mapsto \mathcal{R}^m$ ($m=3$ for RGB image) an input image and $x_{t-1}$ a fixation point at time $t-1$. The next fixation point $x_t$ is determined by sampling the 2D discrete conditional probability $p\left( x|x_{t-1}\right)$ which indicates, for each location of the definition domain $\Omega$, the transition probability between the previous fixation and the current location $x$. The conditional probability $p\left( x|x_{t-1}\right)$ is composed of three terms as described in Eq.~\ref{equ:model}:
\begin{equation}\label{equ:model}
p\left( x|x_{t-1}\right) \propto  p_{BU}(x)p_M(x,t|T)p_B(d(x,x_{t-1}),\phi(x,x_{t-1}))
\end{equation}
where,
\begin{itemize}
\item $p_{BU}(x)$ represents the input 2D bottom-up saliency map. This saliency map is computed by a \textit{traditionnal} saliency model, or by combining the results of several saliency models~\cite{LeMeur2014}.
\item $p_M(x,t|T)$ represents the memory state of the location $x$ at time $t$, according to the $T$ past fixations. This time-dependent term simulates the inhibition of return and indicates the probability to refixate a given location. As described in~\cite{LeMeur2015}, $p_M(x,t|T)$ is composed of two operations: one for inhibiting the current attended location in order to favor the scene exploration. At the opposite, the second term allows to recover the initial salience of the previous attended locations, favoring the re-fixation. An attended location requires $T$ fixations before recovering the integrality of its salience.
\item $p_B(d,\phi)$ represents the probability to observe a saccade of amplitude $d$ and orientation $\phi$. The saccade amplitude $d$, expressed in degree of visual angle, is the Euclidean distance between two consecutive fixation points $x_t$ and $x_{t-1}$. The saccade orientation $\phi$ is the angle, expressed in degree, between these two consecutive fixation points. The joint probability of saccade amplitudes and orientations is learned from actual eye tracking data, by using kernel density estimation~\cite{Silverman1986}. This representation implicitly encompasses gaze biases, which reflect the main tendencies of observers looking at well-defined stimuli. The joint probability is also content-dependent~\cite{LeMeur2016}, indicating that our visual strategy depends on the stimulus displayed on screen (see also supplementary material). By choosing the most relevant joint probability with respect to the displayed scene,  the saccadic model can be fine-tuned for reproducing a specific visual behavior. This is one major difference between saccadic model and  \textit{traditional} saliency models. In  Section~\ref{sec:Helo}, we will see that the joint distribution of saccade amplitudes and orientations is a good candidate for representing the differences in visual deployment that exist between infants and adults.
\end{itemize}

When the three terms of the conditional probability $p\left( x|x_{t-1}\right)$ are known for all sites of the definition domain $\Omega$, the next fixation point can be inferred. One obvious solution would be to consider the maximum a posteriori solution, also called the \textit{Bayesian ideal searcher} in~\cite{Najemnik2009}. However, this solution is deterministic and fails to represent uncertainty about visual perception and perceptual interpretations~\cite{Gershman2012}. Another way to model trial-to-trial variability, or in our context the dispersion between observers, is to assume a stochastic rule for choosing the next fixation point. In~\cite{LeMeur2015}, a set of $N_c$ samples is drawn from the conditional probability $p\left( x|x_{t-1}\right)$. The next fixation point is selected as being the sample having the highest bottom-up salience. This implementation is close to the one proposed in~\cite{Engbert2015spatial}. This form of stochastic selection is also known as Luce's choice rule~\cite{Luce2005individual}. It is important to underline that the number of samples drawn from the conditional probability controls the amount of dispersion between observers. A high number of samples (or candidates)  would reduce the dispersion between observers. In the extreme case, where $N_c$ tends to infinity, the inference of the next fixation point becomes deterministic and strongly similar to the \textit{Bayesian ideal searcher}. At the opposite, when $N_c$ is equal to 1, the amount of randomness is maximal providing the highest dispersion between observers. 

This sampling strategy is obviously sub-optimal because the next fixation point is not necessarily the point having the highest probability to be attended. However, this strategy akin to probability matching~\cite{Wozny2010probability} has been reported to be used by humans in a variety of cognitive tasks~\cite{Gaissmaier2008smart,West2003probability}.

In the next section, we show how this framework is able to capture and implement the specificities of gaze behaviour across the development of the visual system.

\begin{figure*}[!t]
\centering
\subfigure{\includegraphics[width=0.19\linewidth]{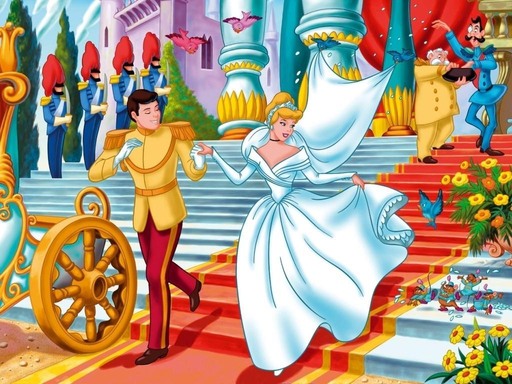}}
\subfigure{\includegraphics[width=0.19\linewidth]{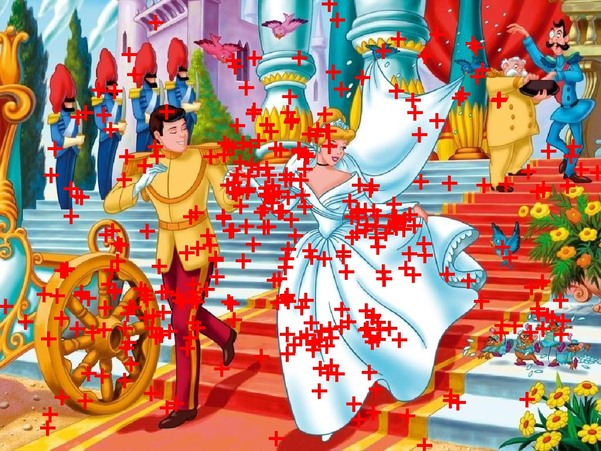}}
\subfigure{\includegraphics[width=0.19\linewidth]{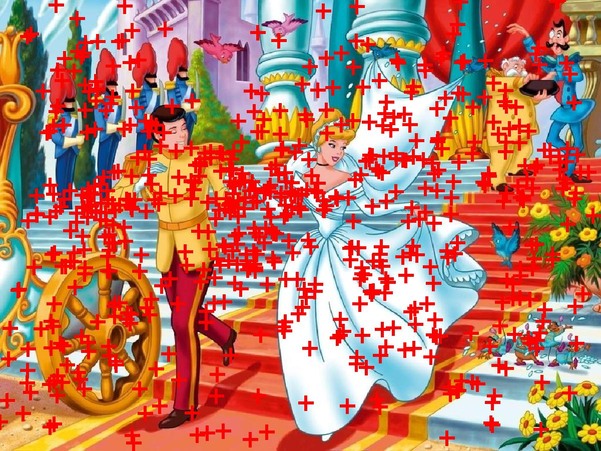}}
\subfigure{\includegraphics[width=0.19\linewidth]{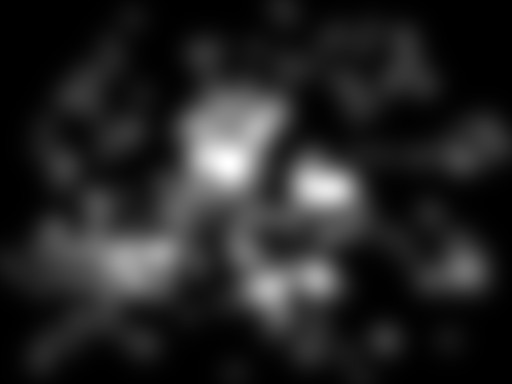}}
\subfigure{\includegraphics[width=0.19\linewidth]{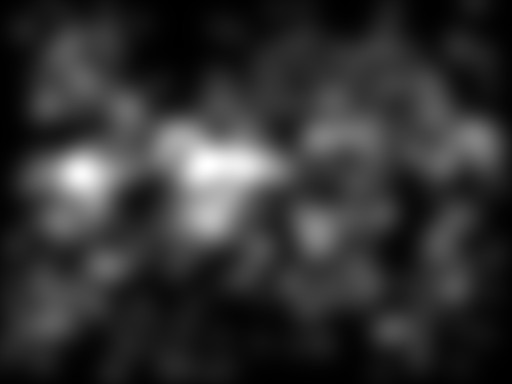}}
\setcounter{subfigure}{0}
\subfigure[]{\includegraphics[width=0.19\linewidth]{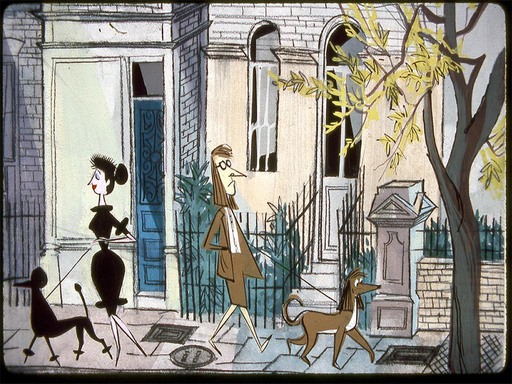}}
\subfigure[]{\includegraphics[width=0.19\linewidth]{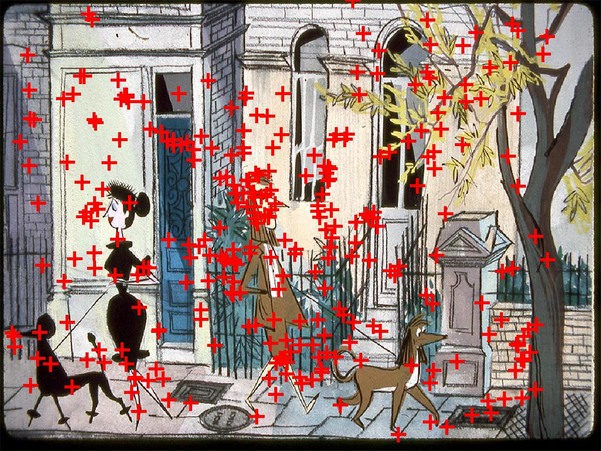}}
\subfigure[]{\includegraphics[width=0.19\linewidth]{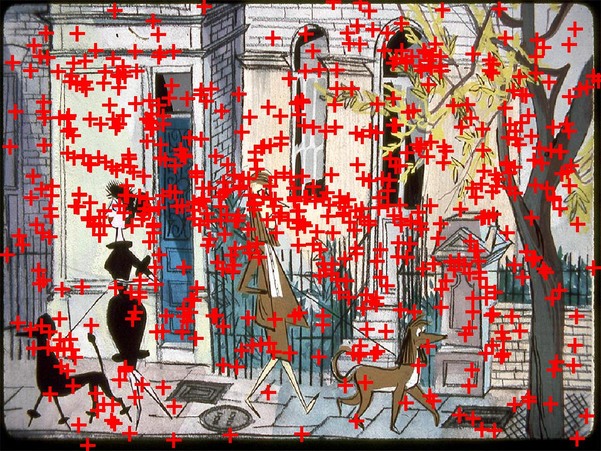}}
\subfigure[]{\includegraphics[width=0.19\linewidth]{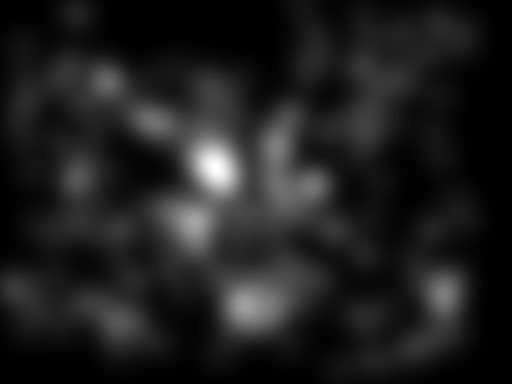}}
\subfigure[]{\includegraphics[width=0.19\linewidth]{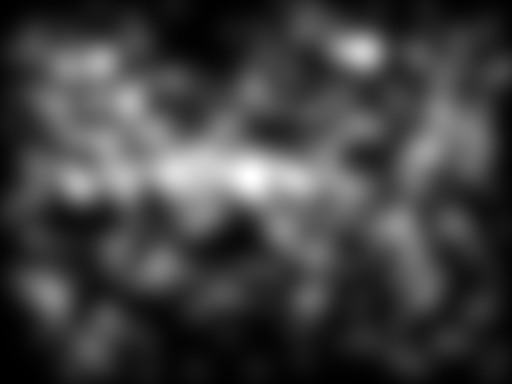}}
\caption{(a) Original stimulus; (b) and (c) represent fixation maps (red crosses indicate fixation) for 2 year-old and adult group, respectively; (d) and (e) represent the actual saliency maps for  2 year-old and adults groups, respectively. }
\label{fig:etData}
\end{figure*}
 
%****************************************************
\section{Eye movements from childhood to adulthood}\label{sec:Helo}
%****************************************************
In this section, we analyzed an eye tracking data collected from observers of a wide range of ages. The main purpose was to investigate whether the joint distribution of saccade orientations and amplitudes learned from this raw eye tracking data is able to capture the gaze biases of different age groups. We already know that aging has an impact on the way we deploy our visual attention~\cite{Luna2008development,Aring2007visual}. If we succeed in quantitatively measuring the influence of development, the saccadic model described in the previous section could be tuned to replicate the gaze behavior of a specific age group.
 
\subsection{Maturation of eye-movements}\label{subsec:maturation}
The visual system at birth is limited but develops rapidly during the first years of life and continues to improve through adolescence. Helo et al.~\cite{Helo2014} give evidence of age-related differences in viewing patterns during free-viewing natural scene perception. Fixation durations decrease with age while saccades turn out to be shorter when comparing children with adults. Materials and methods of this eye-tracking experiment are briefly summarized below.

\subsubsection{Participants} 
A total of 101 subjects participated in the experiments, including 23 adults and 78 children. These subjects were divided into 5 groups: 2 year-old group, 4-6 year-old group, 6-8 year-old group, 8-10 year-old group and adults group.
Participants were instructed to explore the images. The 4-10 year-old and the adults were instructed to perform a recognition test to determine whether an image segment presented at the center of the screen was part of the previous stimulus (more details on experimental design is available in~\cite{Helo2014}). 
\subsubsection{Stimuli} 
Thirty color pictures taken from children books, as illustrated in~\figurename~\ref{fig:etData} (a), are displayed for 10s. A drift correction is performed before each stimulus. The viewing distance is 60 cm. One degree of visual angle represents 28 pixels. For all the  results reported in this paper, the first fixation has been removed.

\subsubsection{Saliency map and center bias}
\begin{figure*}[!t]
\centering
\subfigure[2 year-old]{\includegraphics[width=0.23\linewidth]{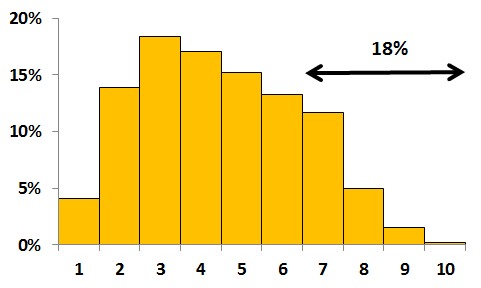}}
\subfigure[4-6 y.o.]{\includegraphics[width=0.23\linewidth]{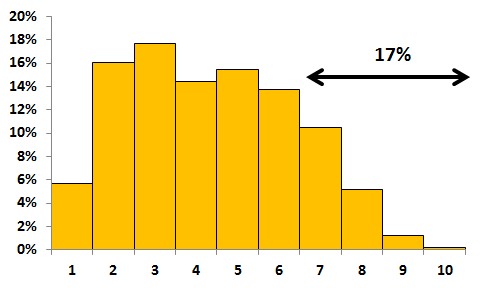}}
\subfigure[6-10 y.o.]{\includegraphics[width=0.23\linewidth]{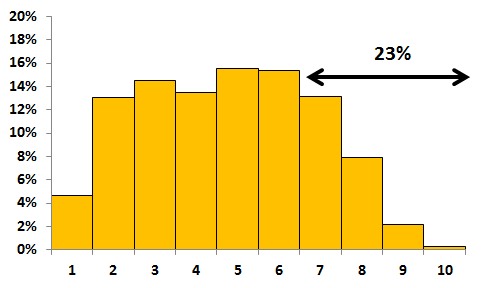}}
\subfigure[Adults]{\includegraphics[width=0.23\linewidth]{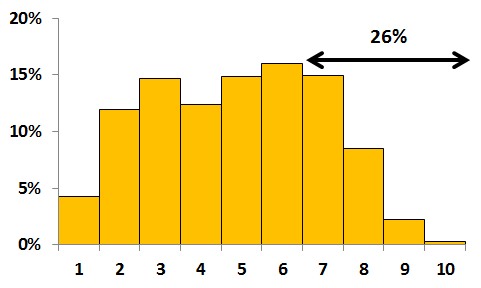}}
\caption{Distribution of visual fixations in function of the distance from the center and distributed into 10 crowns, numbered from 1 to 10 (1 is the centered crown). The y-axis represents a percentage of visual fixations.}
\label{fig:centerBias}
\end{figure*}

\figurename~\ref{fig:etData} illustrates fixation maps and saliency maps computed from eye tracking data of 2 year-old and adults groups. The saliency map is classically computed by convolving actual eye positions with a 2D Gaussian function which approximates the central part of the retina, i.e. the fovea~\cite{Wooding2002fixation}. The standard deviation is set to 28 pixels representing one degree of visual angle~\cite{LeMeur2013}. We observed that adults tend to explore much more the visual scene than 2 year-old children. In addition, the center bias is more important for the 2 year-old group than for the adult group. We quantify this trend by computing the ratio of fixations falling within centered crowns. For this purpose, a set of 10 concentric circles is used. The radius of each circle represents  10\%, 20\%,...,90\%, 100\% of the distance between the picture center and its top-left corner. The ratio of fixations falling within each crown (difference between two concentric successive circles) to the overall number of fixations is calculated. \figurename~\ref{fig:centerBias} plots these distributions for the four groups. The cumulative percentage of the last 4 crowns indicates that the center bias is more significant for young children than adults (26\% of adults' fixations fall within these crowns, compared to only 18\% for 2 year-old children).

\setcounter{subfigure}{0}

\subsection{Joint distribution of saccade orientations and amplitudes}\label{subsec:joint_pdf}
\begin{figure*}[!t]
\centering
\subfigure{\includegraphics[width=0.325\linewidth]{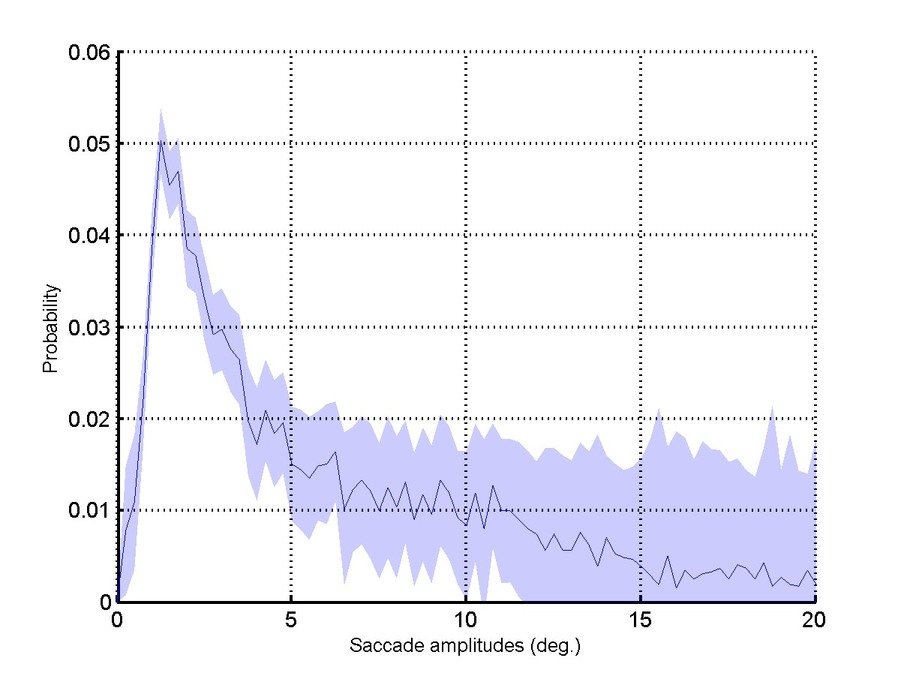}}
\subfigure{\includegraphics[width=0.325\linewidth]{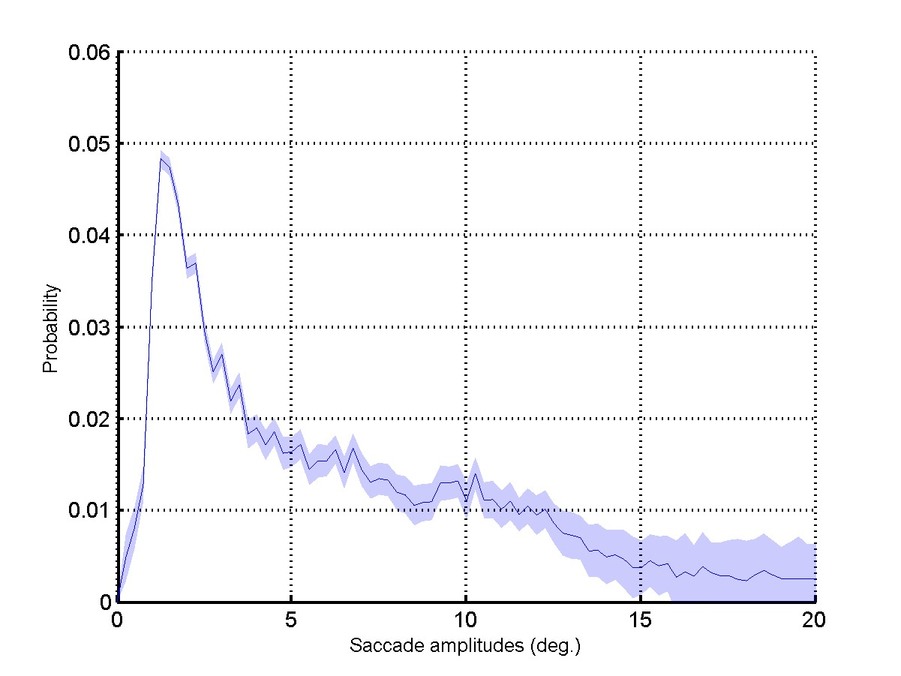}}
%\subfigure{\includegraphics[width=0.24\linewidth]{baby/68yoGT_amplitudeSaccade_A1}}
%\subfigure{\includegraphics[width=0.19\linewidth]{baby/810yoGT_amplitudeSaccade_A1}}
\subfigure{\includegraphics[width=0.325\linewidth]{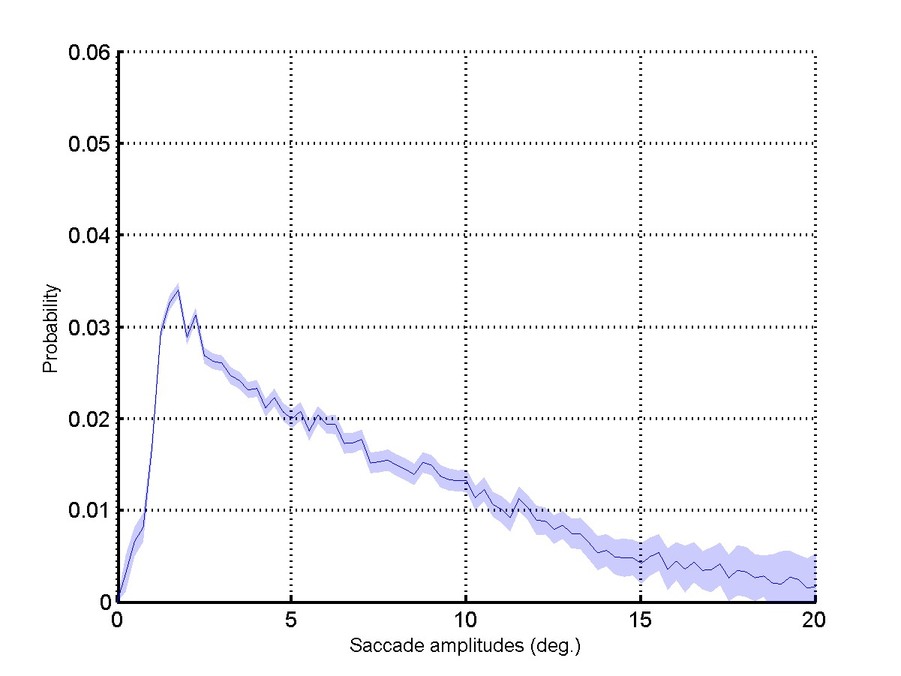}}
\setcounter{subfigure}{0}
\subfigure[2 year-old]{\includegraphics[width=0.325\linewidth]{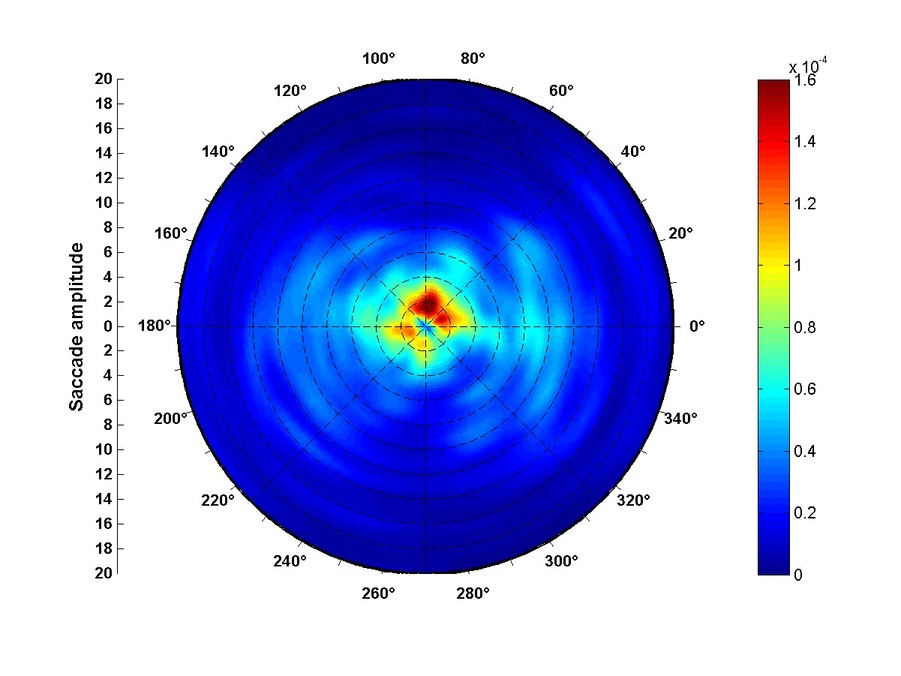}}
\subfigure[4-6 year-old]{\includegraphics[width=0.325\linewidth]{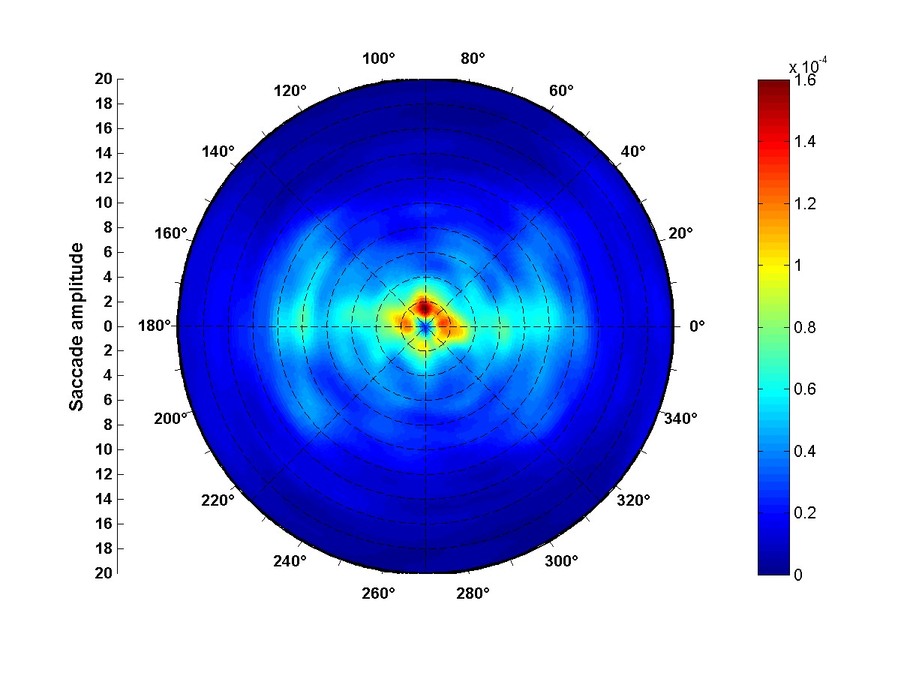}}
%\subfigure[6-8 year-old]{\includegraphics[width=0.24\linewidth]{baby/68yoGT_jointDistribution_A1}}
%\subfigure[8-10 years-old]{\includegraphics[width=0.19\linewidth]{baby/810yoGT_jointDistribution_A1}}
\subfigure[Adults]{\includegraphics[width=0.325\linewidth]{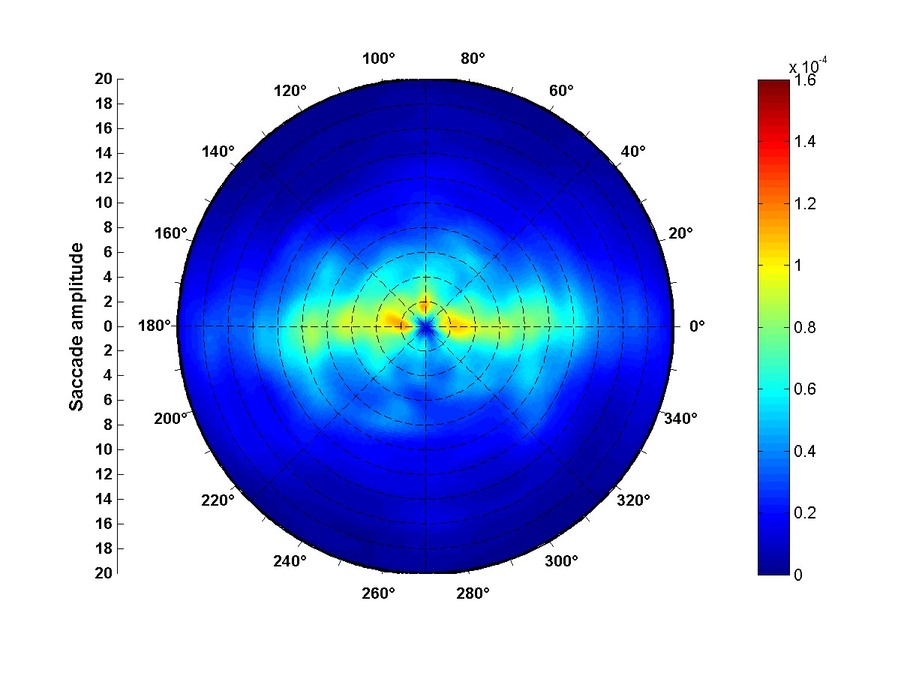}}
\caption{Distribution of saccade amplitudes (top row) and polar plots of joint distribution of saccade amplitudes and orientations (bottom row) for different age groups: (a) 2 year-old group to (d) adult group.  The light blue envelope on top-row curves represents the standard error of the mean, amplified by a factor of 2000. The 6-8 and 8-10 year-old distributions are not displayed for the sake of clarity. They are available in the supplementary materials.}
\label{fig:etJointDistribution}
\end{figure*}

Following the method proposed in~\cite{LeMeur2015}, we estimate the joint probability distribution of saccade amplitudes and orientations $p_B(d,\phi)$ for each age group. This nonparametric distribution is obtained by using a 2D Gaussian kernel density estimation. The two bandwidth parameters are chosen optimally based on the linear diffusion method proposed by~\cite{Botev2010}. The joint probability $p_B(d,\phi)$ is given by:
\begin{equation}\label{equ:kde}
p_B(d,\phi)=\frac{1}{n}\sum_{i}K_h(d-d_i, \phi-\phi_i)
\end{equation}
where $d_i$ and $\phi_i$ are the distance and the angle between each pair of successive fixations respectively. $n$ is the total number of samples and $K_h$ is the two-dimensional Gaussian kernel.
\figurename~\ref{fig:etJointDistribution} shows the joint probability distributions of saccade amplitudes and orientations (bottom row) in a polar plot representation. Radial position indicates saccadic amplitudes expressed in degree of visual angle. The top row of \figurename~\ref{fig:etJointDistribution} shows the marginal probability distributions of saccade amplitudes.

A number of observations can be made: first, eye-movement patterns  change with age. Saccade amplitudes are shorter in the 2 year-old group than in adults group. Saccade amplitudes increase with age. This first observation was consistent with the ones made in~\cite{Helo2014}. Regarding the saccade orientations, we observe a strong horizontal bias in the adult group which is also consistent with previous studies~\cite{Foulsham2008,Tatler2008}. This horizontal bias  can be explained by several factors, such as biomechanical factors, physiological factors and the layout of our natural environment~\cite{vanRenswoude2016infants}. Regarding biomechanical factors, Van Renswoude et al.~\cite{vanRenswoude2016infants} stress the point that horizontal saccades require only the use of one pair of muscles whereas saccades in the other directions requires more than one pair of muscles~\cite{Viviani1977curvature}. This horizontal bias is less obvious for infants, even though it may exist~\cite{vanRenswoude2016infants}. \figurename~\ref{fig:etJointDistribution} (bottom row) also shows that the distribution shape of the 2 year-old group (a) is much more isotropic than the adults' one (d), but with a marked tendency for making upward vertical saccades.

A two-sample two-dimensional Kolmogorov-Smirnov test~\cite{Peacock1983} is performed to test whether  the difference between the joint distributions illustrated in~\figurename~\ref{fig:etJointDistribution} is statistically significant. For two given distributions, we randomly draw 5000 samples and test whether both data sets are drawn from the same  distribution. The tests show significant differences between 2 year-old and 4-6 year-old groups, and between 4-6 year-old and 6-8 year-old groups (all $p < .001$). There is  no difference between 6-8 year-old and 8-10 year-old groups ($p=0.2$). A significant difference is however observed between adults and 8-10 year-old groups ($p = 0.0049$). We reduced the within-group variance by increasing the sample size and merging the 6-8 and 8-10 yo groups together. The resulting group is called the 6-10 yo group.

%This process is repeated 100 times. Except for the distributions of 6-8 and 8-10 years old, we observe a statistically significant difference between all these distributions (see Table~\ref{table:statisticalTest}). Data from 6-8 and 8-10 years old children are then merged together to build a new statistically significant set called 6-10 years old children.

In summary, these results suggest that the joint distribution of saccade amplitudes and orientations is able to grasp gaze behavior differences across age, as well as to reflect important features of development on the visual deployment. 

%\begin{table}
%\caption{p-value of the two-sample two-dimensional Kolmogorov-Smirnov test performed between joint distributions of saccade amplitudes and orientations. The test is performed by drawing 5000 samples from both distributions and test whether they are drawn from the same  distribution. This process is repeated 100 times. We report the minimum, the maximum and the median p-value.} 
%\label{table:statisticalTest}
%\centering %\begin{table*}
%\begin{tabular}{|l|c|}
%\hline
%	\bf Age groups	& \bf p-value ($\lbrace min,median,max\rbrace$)\\
%\hline
%\bf 2 y.o vs 4-6 y.o 	&  $\lbrace 5.9\times 10^{-7}, 5.03\times 10^{-7}, 0.0092\rbrace$ \\
%\bf 4-6 y.o vs 6-8 y.o 	&  $\lbrace 5.9\times 10^{-7}, 3.98\times 10^{-5}, 0.059\rbrace$ \\
%\bf 6-8 y.o vs 8-10 y.o &  $\lbrace 2.15\times 10^{-5}, 0.0967, 0.2\rbrace$ \\
%\bf 8-10 y.o vs adults 		&  $\lbrace 2.25\times 10^{-6}, 0.0108, 0.2\rbrace$ \\
%\hline
%\end{tabular}
%\end{table}

\subsection{To what extent do saliency models predict where infants and adults look?}
In this section, we evaluate the influence of low-level visual features on all fixations. Two saliency models, i.e. GBVS~\cite{Harel2006} and RARE2012~\cite{Riche2013}, are used to compute bottom-up saliency maps. These two models are chosen because of their simplicity and  good performance to predict salient areas~\cite{Riche2013}. Performance is evaluated using the following metrics:
\begin{itemize}
\item The linear correlation coefficient (CC) is computed between two saliency maps. A value of 0 means that the two maps are uncorrelated.
\item The similarity (SIM) is calculated based on the normalized probability distributions of the two maps~\cite{Judd2012}. The similarity is the sum of the minimum values at each point in the distributions. SIM=1 means the distributions are identical whereas SIM=0 means the distributions are completely opposite.
\item The Earth Mover's Distance (EMD) measures the distance between two probability distributions by how much transformation on one distribution would need to undergo to match another (EMD=0 for identical distributions).
\item The metrics called AUC-Judd and AUC-Borji consist in considering the saliency map as a binary classifier to separate positive from negative samples at various thresholds (see~\cite{Riche2013,Borji2013analysis} for a review). A ROC analysis is then performed for computing the Area Under Curve: a score of 1 means that the classification is perfect, whereas a value of 0.5 is the chance level.
\item The normalized scanpath score (NSS) measures the mean value of the normalized saliency map at fixation locations~\cite{Peters2005}. NSS=0 represents the chance level. A high positive value means that fixations fall within salient parts of the scene. 
\end{itemize}
These metrics are complementary: the CC metric is used to compare two saliency maps, SIM and EMD compare two distributions whereas AUC-Judd, AUC-Borji and NSS compare a map with a set of fixations. Readers can refer to~\cite{LeMeur2013,Riche2013,Borji2013analysis} for more details on these metrics.

The performances are given in Table~\ref{table:perfSaliencyModelsSummary}. The results were analyzed using a three-way mixed ANOVA design. Age groups (adults, 6-10 yo, 4-6 yo, or 2 yo) was the between-subjects variable; type of saliency model (GBVS or RARE2012) and type of metric (CC, SIM, EMD, KL, AUC-Judd, AUC-Borji, or NSS) were the within-subjects variables. The three-way ANOVA yielded a significant main effect of age ($F(3,95)=17.55$, $p<.001$), model ($F(1,95=4.87$, $p=0.03$) and metric ($F(6,90)=784.84$, $p<.001$). The $metric\times age$ interaction is significant ($F(18,276) = 8.10$, $p<.001$), as well as the $model\times metric$ interaction ($F(6,90) = 59.91, p < .001$). The $model\times age$ interaction is not significant ($F(3,95) = 2.53, p = 0.062$). Post-hoc Bonferroni comparisons show significant differences between all age groups ($p < .001$), except between adults and 6-10 yo, and between 4-6 yo and 2 yo ($p = 1$). 

\begin{figure}
\centering
    \subfigure{\includegraphics[width=0.45\linewidth]{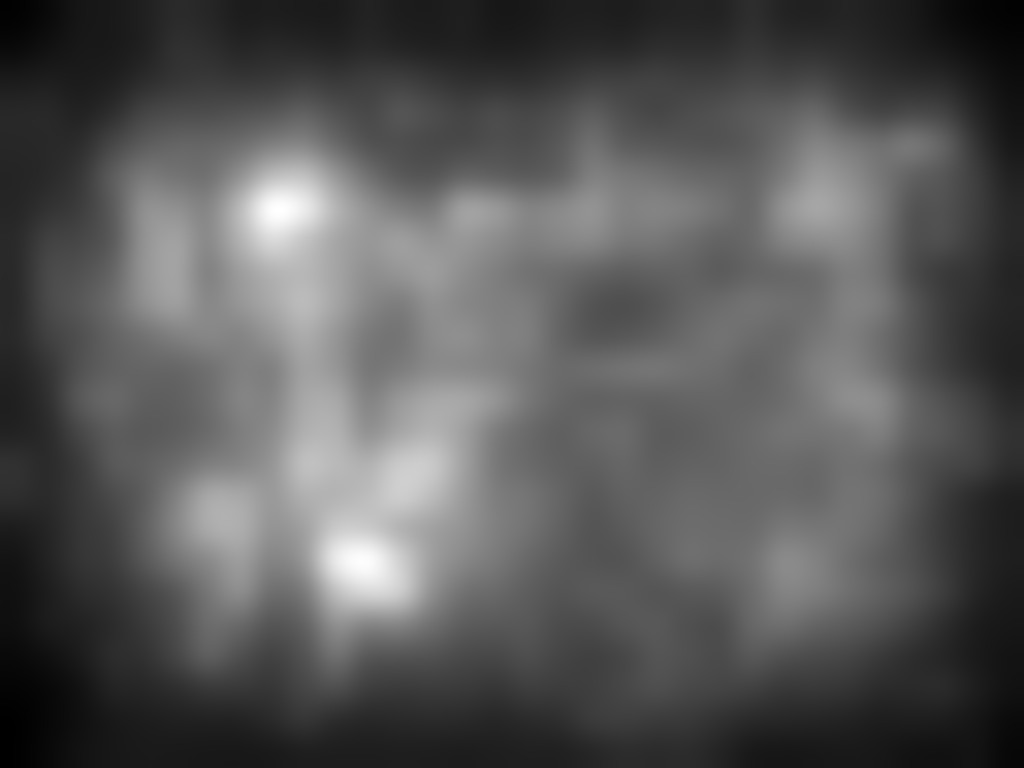}}
    \subfigure{\includegraphics[width=0.45\linewidth]{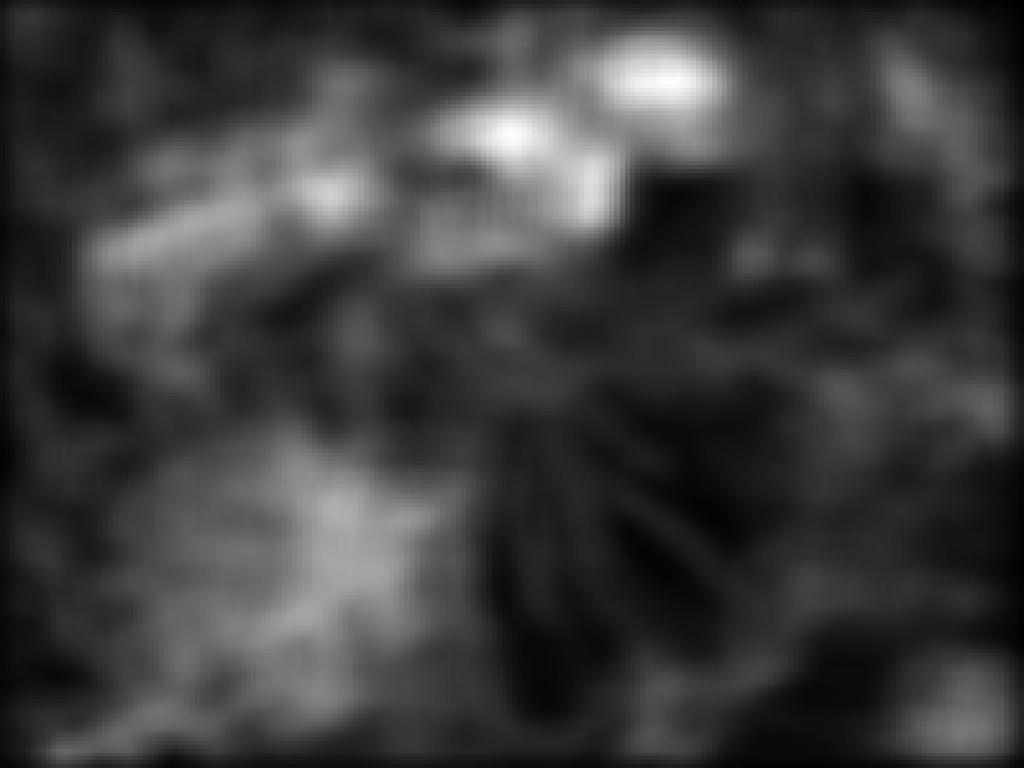}}      
  \captionof{figure}{GBVS (left) and RARE2012 (right) saliency maps for the original image in~\figurename~\ref{fig:etData} (a).}
  \label{fig:predictedSaliencyMap}
\end{figure}

This analysis leads to two main observations. First, the influence of bottom-up factors such as saliency in eye movement behavior is significant for all age groups, but more specifically for GBVS model. Indeed GBVS model significantly outperforms RARE2012 model (paired t-test, $p<<0.01$). According to previous benchmarks of computational models of saliency~\cite{Borji2012,LeMeur2015}, this discrepancy in performance between these two models is unusual. This performance gap might be explained by two major differences between GBVS and RARE2012. %as illustrated by~\figurename~\ref{fig:etData}. 
First, the central bias, while intrinsically taken into account by GBVS, is not considered by RARE2012. As illustrated by~\figurename~\ref{fig:predictedSaliencyMap} (a) and (b), we can observe black stripes all around GBVS saliency map, which may significantly improve the performance of the model~\cite{Bruce2015computational}. Second, RARE2012 maps are much more focused than GBVS ones. GBVS maps being less focused might be an advantage for two reasons. First, the stimuli used in the eye tracking experiments (see~\figurename~\ref{fig:etData}) are very dense, containing several areas of interest. Second, except for the 2 year-old kids, all participants performed a recognition task which might favor  scene exploration, and penalize too clustered saliency maps.
%Finally, we previously shown in~\figurename~\ref{fig:centerBias} that there exist a center bias in the eye tracking dataset we use. 

A second observation is related to the influence of salience in the four age groups. The best match is obtained for the 6-10 year-old group when considering the CC, SIM and EMD metrics, for both saliency models.  For the other three metrics, i.e. AUC-Judd, AUC-Borji and NSS, the best scores are obtained for the 4-6 year-old group.  These results, showing a better match between the predicted saliency maps and eye tracking data for children from the age of 4 to 10 years than for older participants, generally agree with previous findings. Authors in~\cite{Acik2010} showed that bottom-up factors decrease with age while the role of top-down processes increases. However, we observe that the role of bottom-up factors is less important for the 2 year-old group than for children between 4 and 10 year-old, which is not in agreement with~\cite{Helo2014}. Performance, whatever the metric and saliency model, follows an inverted U-shape curve, in which young groups do not systematically outperform the adult group.

%****************************************************
\section{Age-dependent saccadic model}\label{sec:adaptationSaccadicModel}
%****************************************************
In this section we tailor Le Meur's saccadic model to the different age groups, namely, 2, 4-6, 6-10 year-old and adults. We perform three modifications to the purpose of our study. The first modification consists in using a joint probability density function $p_B(d,\phi)$ that has been learned from eye tracking data collected from different age groups, as presented in section~\ref{subsec:joint_pdf}. This prior knowledge represents the viewing tendencies, expressed in this study in terms of saccade amplitudes and orientations, which are common across all observers of a given age. The use of such a prior is fundamental to constrain how we explore scenes and to generate saccade amplitudes and orientations that match those estimated from human eye behavior. 

However, rather than using a unique joint distribution per age group, we use, as suggested in~\cite{LeMeur2016}, a spatially-variant joint distribution. The image is then split into a non-overlapping $3 \times 3$ grid; for each cell in the grid, the joint distribution of saccade amplitudes and saccade orientations is estimated following the procedure detailed in section~\ref{subsec:joint_pdf} (the polar plots of these distributions are given in supplementary materials). This spatially-variant prior is more appropriate for catching important viewing tendencies.  One of the most important priors is the central bias. Indeed, as illustrated in~\cite{LeMeur2016} as well as in the supplementary materials, the joint distributions of saccade amplitudes and orientations located on the frame corners consist of saccades going towards the screen's center. This reflects our tendency to look near the screen's center, irrespective of the visual information at that location.

The second modification is related to the number of samples, $N_c$, which is drawn from the conditional probability $p\left( x|x_{t-1}\right)$. As presented in section~\ref{sec:saccadicModel}, this parameter can be used to tune the amount of randomness in the selection of the next fixation point. A low value results in a high dispersion between observers, and fosters the scene exploration~\cite{Martinez2008stochastic}. A high value would reduce the dispersion.  In our study, we evaluate the performance of the saccadic model for $N_c\in\lbrace1,...,9\rbrace$.  
%The extreme case, when $N_c$ tends to infinity, boils down to a standard winner-take-all scheme~\cite{Itti1998}.

The third modification concerns the selection of the most appropriate candidate among the $N_c$ candidates drawn from the conditional probability $p\left( x|x_{t-1}\right)$. In~\cite{LeMeur2015}, the next fixation point is selected as being the candidate having the highest bottom-up salience. This selection rule is modified to take into account the probability $p_B(.,.)$, the bottom-up saliency $p_{BU}(.)$ and the distance $d$ between the candidate and the previous fixation point. The next fixation point $x^*$ is then selected as 
\begin{equation}
x^*=\arg\max_{s\in\Theta} \frac{p_{BU}(s)\times p_B(d(s,x_{t-1}),\phi(s,x_{t-1}))}{d(s,x_{t-1})}
\end{equation}
where, $\Theta$ is the set of $N_c$ candidates, $x_{t-1}$ is the previous fixation point and $d$ is the Euclidean distance between the candidate $s$ and the previous fixation point. This new rule allows us to favor the candidates that are close to the previous fixation point and featured by both a high probability to be attended and high bottom-up salience.

\begin{table}
\caption{Performance of GBVS, RARE2012 models and saccadic model (using an input saliency map computing from either GBVS or RARE2012). We report the performances of our saccadic model for an optimal $N_c$ value. The best scores are in bold. Detailled performances are given in Table~\ref{table:perfSaliencyModels}.}\label{table:perfSaliencyModelsSummary}
\scriptsize
\centering
  	\begin{tabular}{|l|c|c|c|c|c|c|}
	\hline
	\bf Metrics 			& \bf CC & \bf SIM &   \bf EMD & \bf AUC-Judd & \bf AUC-Borji & \bf NSS\\
	\hline
	\hline
	\multicolumn{7}{|c|}{\bf Adults}\\
	\cline{1-7}
GBVS    	&0.531		&\bf 0.731	&0.868		&\bf 0.644	&\bf 0.634	&0.463\\
Our model   &\bf 0.636  &0.706		&\bf 0.759  &\bf 0.644  &\bf 0.639  &0.561\\
RARE2012	&0.290	&0.638	&1.228	&0.592	&0.575	&0.256\\ 
Our model   &\bf 0.566 &\bf 0.701 &\bf 0.787 &\bf 0.640 &\bf 0.630 &\bf 0.492\\
	\hline
	\multicolumn{7}{|c|}{\bf 6-10 y.o.}\\
	\cline{1-7}
GBVS    	&0.589	&\bf 0.761	&0.776	&\bf 0.661	&\bf 0.640	&0.478	\\
Our model   &\bf 0.686 &0.732 &\bf 0.744 & 0.659 &\bf 0.640 &\bf 0.562\\
RARE2012	&0.328	&0.661	&1.183	&0.607	&0.578	&0.266\\
Our model   &\bf 0.617 &\bf 0.717 &\bf 0.792 &\bf 0.648 &\bf 0.630 &\bf 0.505\\
	\hline
	\multicolumn{7}{|c|}{\bf 4-6 y.o.}\\
	\cline{1-7}
GBVS    	&0.544	&\bf 0.691	&1.052	&\bf 0.690	&0.675	&0.597	\\
Our model   &\bf 0.673 & 0.690 &\bf 0.806 &0.688 &\bf 0.681 &\bf 0.745\\
RARE2012	&0.275	&0.592	&1.464	&0.614	&0.587	&0.296\\
Our model   &\bf 0.602 &\bf 0.683 &\bf 0.900 &\bf 0.678 &\bf 0.672 &\bf 0.661\\
	\hline
	\multicolumn{7}{|c|}{\bf 2 y.o.}\\
	\cline{1-7}
GBVS    	&0.501	&\bf 0.662	&1.071	&\bf 0.674	&0.667	& 0.570	\\
Our model   &\bf 0.579 &0.659 &\bf 0.906 &\bf 0.674 &\bf 0.671 &\bf 0.666\\
RARE2012	&0.264	& 0.578	&1.431	&0.601	&0.579	&0.292\\
Our model   &\bf 0.517 &\bf 0.639 &\bf 1.014 &\bf 0.662 &\bf 0.653 &\bf 0.585\\
	\hline
	\end{tabular}  
\end{table}

\begin{table*}
\caption{Performance of GBVS, RARE2012 models and saccadic model. The best scores are in bold. A dagger, i.e. ${}^{\dagger}$, is added when there is a statistically significant difference (paired t-test, $p<0.05$) between GBVS (resp. RARE2012) and GBVS-based saccadic model (resp. RARE2012-based saccadic model). The bullet, i.e. ${}^{\bullet}$, indicates the scores that are NOT statistically significant: the paired t-test is performed in this case between the highest score (in bold) and other scores obtained by varying $N_c$. On the last rows, \textit{NSV dist.} means Non Spatially Variant joint distribution and SVdist2yo means Spatially Variant joint distribution of 2 y.o. group.}\label{table:perfSaliencyModels}
\centering
  	\begin{tabular}{|l|c|c|c|c|c|c||c|c|c|c|c|c|}
	\hline
	\bf Metrics 			& \bf CC & \bf SIM &   \bf EMD & \bf AUC-Judd & \bf AUC-Borji & \bf NSS	&\bf  CC &\bf  SIM &\bf   EMD & \bf AUC-Judd & \bf AUC-Borji & \bf NSS\\
	\hline
	\hline
	\multicolumn{13}{|c|}{\bf Adults}\\
	\cline{2-13}
		& \multicolumn{6}{|c||}{\bf GBVS model~\cite{Harel2006}} & \multicolumn{6}{|c|}{\bf RARE2012 model~\cite{Riche2013}} \\
	\cline{2-13}
	    &0.531	&\bf 0.731${}^{\dagger}$	& 0.868	&\bf 0.644	&\bf 0.634	&0.463	&0.290	&0.638	&1.228	&0.592	&0.575	&0.256\\ %${}^{\dagger}$
	\cline{2-13}
		& \multicolumn{6}{|c||}{\bf GBVS-based saccadic model} & \multicolumn{6}{|c|}{\bf RARE2012-based saccadic model} \\
	\hline					
	Nc=1	&0.471		&0.706	&\bf 0.759${}^{\dagger}$	&0.621	&0.615		&0.416		&0.459		&\bf 0.701${}^{\dagger}$	&\bf 0.787${}^{\dagger}$	&0.617	&0.610	&0.402\\
	Nc=2	&0.580		&0.704	&0.821${}^{\bullet}$		&0.641${}^{\bullet}$		&0.633 		&0.507 		&\bf 0.566${}^{\dagger}$		&0.698${}^{\bullet}$ 	&0.824${}^{\bullet}$ &\bf 0.640${}^{\dagger}$ &\bf  0.630${}^{\dagger}$ &\bf 0.492${}^{\dagger}$\\ 	 
	Nc=3	&0.619${}^{\bullet}$		&0.686	&1.029		&\bf 0.644	&\bf 0.639	&0.541${}^{\bullet}$		&0.561${}^{\bullet}$	&0.671		&1.094	&0.639${}^{\bullet}$	&0.627${}^{\bullet}$	&\bf 0.492${}^{\dagger}$\\
	Nc=4	&\bf 0.636${}^{\dagger}$	&0.676		&1.099		&0.643${}^{\bullet}$		&0.635${}^{\bullet}$		&\bf 0.561${}^{\dagger}$		&0.537${}^{\bullet}$		&0.644		&1.250	&0.630	&0.619	&0.467${}^{\bullet}$\\
	Nc=5    &0.617${}^{\bullet}$		&0.651	&1.244		&0.634		&0.629		&0.540${}^{\dagger}$		&0.524		&0.626		&1.342	&0.626	&0.613	&0.460${}^{\bullet}$\\
	Nc=6	&0.615${}^{\bullet}$		&0.485	&1.285		&0.637		&0.628		&0.537		&0.503		&0.605		&1.474	&0.616	&0.605	&0.441${}^{\bullet}$\\
	Nc=7    &0.620		&0.638	&1.368		&0.635		&0.629		&0.543		&0.500		&0.599		&1.442	&0.619	&0.605	&0.441${}^{\bullet}$\\
	Nc=9    &0.606		&0.629	&1.429		&0.634		&0.625		&0.535		&0.483		&0.581		&1.585	&0.611	&0.600	&0.424\\
	\hline
	\multicolumn{13}{|c|}{\bf 6-10 y.o.}\\
	\cline{2-13}
		& \multicolumn{6}{|c||}{\bf GBVS model~\cite{Harel2006}} & \multicolumn{6}{|c|}{\bf RARE2012 model~\cite{Riche2013}} \\
	\cline{2-13}
	    &0.589	&\bf 0.761	&0.776	&\bf 0.661	&\bf 0.640	&0.478	&0.328	&0.661	&1.183	&0.607	&0.578	&0.266\\
	\cline{2-13}
		& \multicolumn{6}{|c||}{\bf GBVS-based saccadic model} & \multicolumn{6}{|c|}{\bf RARE2012-based saccadic model} \\
	\hline					
	Nc=1	&0.479		&0.716	&\bf 0.744	&0.620	&0.608		&0.390		&0.448		&0.716${}^{\bullet}$		&\bf 0.792${}^{\dagger}$	&0.621		&0.602	&0.367\\
	Nc=2	&0.649		&0.732	&0.740${}^{\bullet}$		&0.656	&\bf 0.640	&0.532		&0.588		&\bf 0.717${}^{\dagger}$		&0.816${}^{\bullet}$		&\bf 0.648${}^{\dagger}$		&0.629${}^{\bullet}$	&0.477\\
	Nc=3	&0.666		&0.707	&0.954		&0.658	&\bf 0.640	&0.545		&\bf 0.617${}^{\dagger}$	&0.698	&0.982		&\bf 0.648${}^{\dagger}$	&\bf 0.630${}^{\dagger}$	&\bf 0.505${}^{\dagger}$\\
	Nc=4	&\bf 0.686${}^{\dagger}$	&0.698	&1.011		&0.659	&\bf 0.640	&\bf 0.562${}^{\dagger}$ 	&0.583		&0.658		&1.277		&0.642${}^{\bullet}$		&0.621	&0.475\\
	Nc=5	&0.655		&0.675	&1.162		&0.655	&0.635		&0.535		&0.569		&0.639		&1.330		&0.639		&0.616	&0.463\\
	Nc=6	&0.667		&0.661	&1.258		&0.646	&0.631		&0.545		&0.560		&0.625		&1.355		&0.637		&0.613	&0.455\\
	Nc=7	&0.667		&0.655	&1.283		&0.647	&0.631		&0.546		&0.547		&0.608		&1.494		&0.630		&0.607	&0.448\\
	Nc=9	&0.660		&0.647	&1.361		&0.646	&0.629		&0.544		&0.525		&0.586		&1.655		&0.627		&0.600	&0.429\\
	\hline
	\multicolumn{13}{|c|}{\bf 4-6 y.o.}\\
	\cline{2-13}
		& \multicolumn{6}{|c||}{\bf GBVS model~\cite{Harel2006}} & \multicolumn{6}{|c|}{\bf RARE2012 model~\cite{Riche2013}} \\
	\cline{2-13}
	    &0.544	&\bf 0.691	&1.052	&\bf 0.690	&0.675	&0.597	&0.275	&0.592	&1.464	&0.614	&0.587	&0.296\\
	\cline{2-13}
		& \multicolumn{6}{|c||}{\bf GBVS-based saccadic model} & \multicolumn{6}{|c|}{\bf RARE2012-based saccadic model} \\
	\hline					
	Nc=1 &0.506		&0.666	&0.965		&0.657	&0.651		&0.556		&0.494		&0.664	&0.987${}^{\bullet}$	&0.652	&0.647	&0.541	\\
	Nc=2 &0.630		&0.690	&\bf 0.806${}^{\dagger}$	&0.684	&0.680${}^{\bullet}$		&0.692		&\bf 0.602${}^{\dagger}$		&\bf 0.683${}^{\dagger}$	&\bf 0.900${}^{\dagger}$	&\bf 0.678${}^{\dagger}$		&\bf 0.672${}^{\dagger}$	&\bf 0.661${}^{\dagger}$\\
	Nc=3 &0.660${}^{\bullet}$		&0.687	&0.940		&0.688	&\bf 0.681	&0.730${}^{\bullet}$		&0.592${}^{\bullet}$	&0.662	&1.037	&0.677${}^{\bullet}$	&0.667	&0.651${}^{\bullet}$	\\
	Nc=4 &\bf 0.673${}^{\dagger}$	&0.669	&1.020		&0.683	&0.675${}^{\bullet}$		&0.744${}^{\bullet}$		&0.569		&0.632	&1.216	&0.671		&0.655		&0.624\\
	Nc=5 &\bf 0.673${}^{\dagger}$	&0.663	&1.108		&0.685	&0.675${}^{\bullet}$		&\bf 0.745${}^{\dagger}$	&0.552		&0.619	&1.295	&0.667	&0.649	&0.605	\\
	Nc=6 &0.663${}^{\bullet}$		&0.651	&1.131		&0.680	&0.669		&0.737${}^{\bullet}$		&0.546		&0.605	&1.366	&0.660	&0.644	&0.596\\
	Nc=7 &0.658${}^{\bullet}$		&0.645	&1.221		&0.677	&0.670		&0.730${}^{\bullet}$		&0.539		&0.598	&1.393	&0.663	&0.641	&0.595	\\
	Nc=9 &0.650		&0.632	&1.238		&0.675	&0.665		&0.720		&0.504		&0.565	&1.508	&0.651	&0.628	&0.550	\\
	\hline
	\multicolumn{13}{|c|}{\bf 2 y.o.}\\
	\cline{2-13}
		& \multicolumn{6}{|c||}{\bf GBVS model~\cite{Harel2006}} & \multicolumn{6}{|c|}{\bf RARE2012 model~\cite{Riche2013}} \\
	\cline{2-13}
	    &0.501	&\bf 0.662	&1.071	&\bf 0.674	&0.667	& 0.570	&0.264	& 0.578	&1.431	&0.601	&0.579	&0.292\\
	\cline{2-13}
		& \multicolumn{6}{|c||}{\bf GBVS-based saccadic model} & \multicolumn{6}{|c|}{\bf RARE2012-based saccadic model} \\
	\hline					
	Nc=1	&0.385		&0.624	&1.157		&0.628		&0.622		&0.445		&0.413		&0.627	&1.136			&0.628	&0.629	&0.468	\\
	Nc=2	&0.556${}^{\bullet}$		&0.659	&\bf 0.906${}^{\dagger}$	&0.670		&0.670${}^{\bullet}$		&0.640		&0.492${}^{\bullet}$		&\bf 0.639	&\bf 1.015${}^{\dagger}$			&0.660${}^{\bullet}$	&\bf 0.653${}^{\dagger}$	&0.556${}^{\bullet}$\\
	Nc=3	&0.577${}^{\bullet}$		&0.653	&1.015${}^{\bullet}$		&\bf 0.674	&\bf 0.671	&0.661		&\bf 0.517${}^{\dagger}$	&0.632	&1.081${}^{\bullet}$	&\bf 0.662${}^{\dagger}$	&\bf 0.653${}^{\dagger}$	&\bf 0.585${}^{\dagger}$	\\
	Nc=4	&0.575${}^{\bullet}$		&0.634	&1.129		&0.667		&0.663${}^{\bullet}$		&0.665${}^{\bullet}$		&0.475		&0.599 	&1.304			&0.653${}^{\bullet}$	&0.636	&0.536\\
	Nc=5	&0.575${}^{\bullet}$		&0.635	&1.130		&0.668		&0.663${}^{\bullet}$		&\bf 0.666${}^{\dagger}$	&0.476		&0.599	&1.305			&0.653${}^{\bullet}$	&0.642	&0.536	\\
	Nc=6	&0.571${}^{\bullet}$		&0.628	&1.166		&0.670		&0.662${}^{\bullet}$		&0.657${}^{\bullet}$		&0.468		&0.582	&1.419			&0.647	&0.631	&0.532\\
	Nc=7	&\bf 0.579${}^{\dagger}$	&0.629	&1.146		&0.668		&0.664${}^{\bullet}$		&0.665${}^{\bullet}$		&0.439		&0.566	&1.486			&0.646	&0.625	&0.497	\\
	Nc=9	&0.569${}^{\bullet}$		&0.616	&1.321		&0.658		&0.655		&0.661${}^{\bullet}$		&0.441		&0.557	&1.563			&0.639	&0.620	&0.498	\\
	\hline
	\hline	
	\multicolumn{13}{|c|}{\bf Influence of joint distribution (Nc=4, Adults) }\\
	\cline{2-13}
%		& \multicolumn{6}{|c||}{\bf GBVS model~\cite{Harel2006}} & \multicolumn{6}{|c|}{\bf RARE2012 model~\cite{Riche2013}} \\
%	\cline{2-13}
%	    &0.531	&\bf 0.731${}^{\dagger}$	& 0.868	&\bf 0.644	&\bf 0.634	&0.463	&0.290	&0.638	&1.228	&0.592	&0.575	&0.256\\ %${}^{\dagger}$
	\cline{2-13}
		& \multicolumn{6}{|c||}{\bf GBVS-based saccadic model} & \multicolumn{6}{|c|}{\bf RARE2012-based saccadic model} \\
	\hline					
	\small{SVdist2yo}		&0.576		&0.647	&1.285		&0.632		&0.623		&0.504		&0.507		&0.633	&1.297			&0.623	&0.613	&0.446 \\	 
	\tiny{$p_B(d,\phi)=1$}	&0.497		&0.667	&1.106		&0.631		&0.621		&0.430		&0.362		&0.630	&1.323			&0.603	&0.589	&0.318	\\
	\small{NSV dist.}		&0.531		&0.685	&1.072		&0.631		&0.624		&0.462		&0.405		&0.647	&1.244			&0.608	&0.597	&0.354	\\
	\hline
	\end{tabular}  
\end{table*}

%****************************************************
\section{Performances}\label{sec:perf}
%****************************************************

First, we evaluate the extent to which the predicted fixations fall within salient areas.  Second, we test the plausibility of the generated scanpaths with respect to the actual scanpaths of the four age groups. Third, we evaluate the benefit to use dedicated age-dependent distributions of saccade amplitudes and orientations. Fourth, the influence of the parameter $N_c$ is analyzed.

To perform this evaluation, we proceed as follows: for each image of the dataset and for each age group, we generate 20 scanpaths, each composed of 15 fixations. The first fixation is randomly chosen. The input saliency map, i.e. the term $p_{BU}$ in equation~\ref{equ:model}, is computed using either the GBVS or RARE2012 model. From the generated scanpaths, a saliency map is computed by following the classical procedure, as described in section~\ref{subsec:maturation} (see also~\cite{Wooding2002fixation,LeMeur2013}). These maps are called scanpath-based saliency maps.
  
\subsection{Prediction of salient areas}
Table~\ref{table:perfSaliencyModelsSummary} and~\ref{table:perfSaliencyModels} present the similarity degree between scanpath-based saliency map and the ground truth (i.e. either human saliency map or eye tracking data).  Table~\ref{table:perfSaliencyModelsSummary} provides the performance of our saccadic model for an optimal value of $N_c$. We observe that our model significantly outperforms RARE2012 model, whatever age groups and metrics. Compared to GBVS model, our model performs better according to 4 metrics. A thorough statistical analysis is performed from the detailed scores given in Table~\ref{table:perfSaliencyModels}. The results were analyzed using 2 three-way mixed ANOVA designs. 

The first one uses age groups (adults, 6-10 yo, 4-6 yo, or 2 yo) as the between-subjects variable, type of saliency model (GBVS or GBVS-based saccadic model) and type of metric (CC, SIM, EMD, AUC-Judd, AUC-Borji, or NSS) as the within-subjects variables. For each metric and age group, we used the $N_c$ value that led to the best result. The three-way ANOVA yielded a significant main effect of age ($F(3, 95) = 15.31$, $p < .001$), model ($F(1, 95) = 8.056$, $p = 0.006$) and metric ($F(5,91)=110.50$, $p < .001$). The $metric\times age$ interaction is significant ($F(15,279) = 9.035$, $p < .001$), as well as the $model\times metric$ interaction ($F(5,91) = 52.32$, $p < .001$). The $model\times age$ interaction is not significant ($F(3,95) = 1.25$, $p = 0.29$). Post-hoc Bonferroni comparisons show significant differences between all age groups (all $p <.001$, except between 2 yo and 6-10 yo where $p = 0.035$), except between 4-6 yo and 2 yo ($p = 0.49$). 

The second ANOVA analysis uses age groups (adults, 6-10 yo, 4-6 yo, or 2 yo) as the between-subjects variable, type of saliency model (RARE or RARE-based saccadic model) and type of metric (CC, SIM, EMD, AUC-Judd, AUC-Borji, or NSS) as the within-subjects variables. For each metric and age group, we used the $N_c$ value that led to the best result.
The three-way ANOVA yielded a significant main effect of age ($F(3, 95) = 6.81$, $p < .001$), model ($F(1, 95) = 67.92$, $p < 0.001$) and metric ($F(5,91)=262.45$, $p < .001$). The $metric\times age$ interaction is significant ($F(15,279) = 7.43$, $p < .001$), as well as the $model\times metric$ interaction ($F(5,91) = 96.143$, $p < .001$). The $model\times age$ interaction is not significant ($F(3,95) = 0.62$, $p = 0.60$). Post-hoc Bonferroni comparisons show significant differences between adults and 4-6 yo ($p < .001$), marginal differences between adults and 2 yo ($p = 0.086$) and no difference between adults and 6-10 yo ($p = 1$). There is a significant difference between 6-10 yo and 4-6 yo ($p = 0.01$) but not between 6-10 yo and 2 yo ($p = 0.57$). There is no significant difference between 2 yo and 4-6 yo ($p = 1$).

In summary, as shown in Table~\ref{table:perfSaliencyModels}, the proposed saccadic model performs better than GBVS and RARE2012 models.  When the input saliency map of the saccadic model is the saliency map computed by RARE2012, the saccadic model significantly outperforms RARE2012 for all considered similarity metrics. These results are given on the right hand-side of Table~\ref{table:perfSaliencyModels}.  We draw a similar conclusion for the CC, EMD and NSS metrics when the GBVS model is used to compute the input saliency map. Concerning the SIM, AUC-Judd and AUC-Borji metrics, the performances of the GBVS-based saccadic model are similar to GBVS model.

The proposed model performs well when $N_c$ is in between 3 and 7, for all age groups. This shows a reasonable flexibility with the choice of the $N_c$ parameter. As discussed in the next section, the parameter $N_c$ appears to be much more important when it comes to generate plausible visual scanpaths.

\subsection{Are visual scanpaths plausible?}
\begin{figure*}
\centering  
    \subfigure{\includegraphics[width=0.325\linewidth]{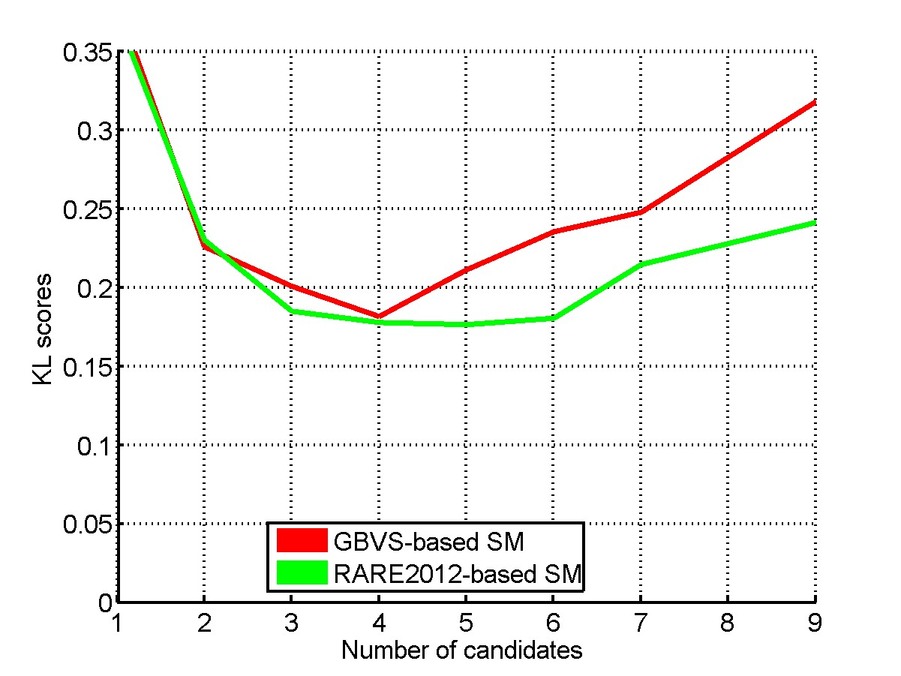}}    
    \subfigure{\includegraphics[width=0.325\linewidth]{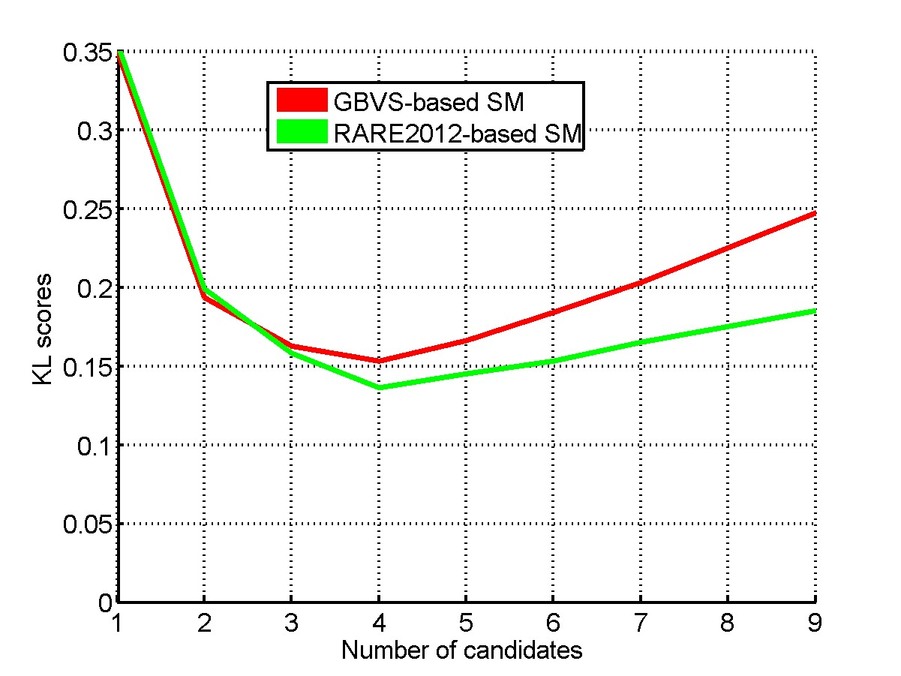}}    
    \subfigure{\includegraphics[width=0.325\linewidth]{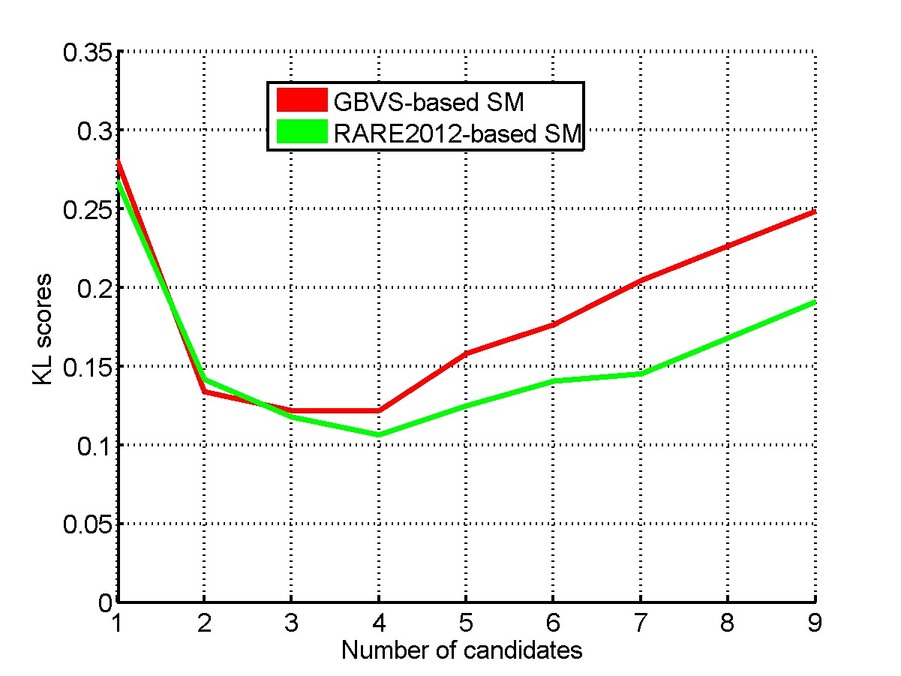}}
    \setcounter{subfigure}{0}
    \subfigure[2 years-old]{\includegraphics[width=0.325\linewidth]{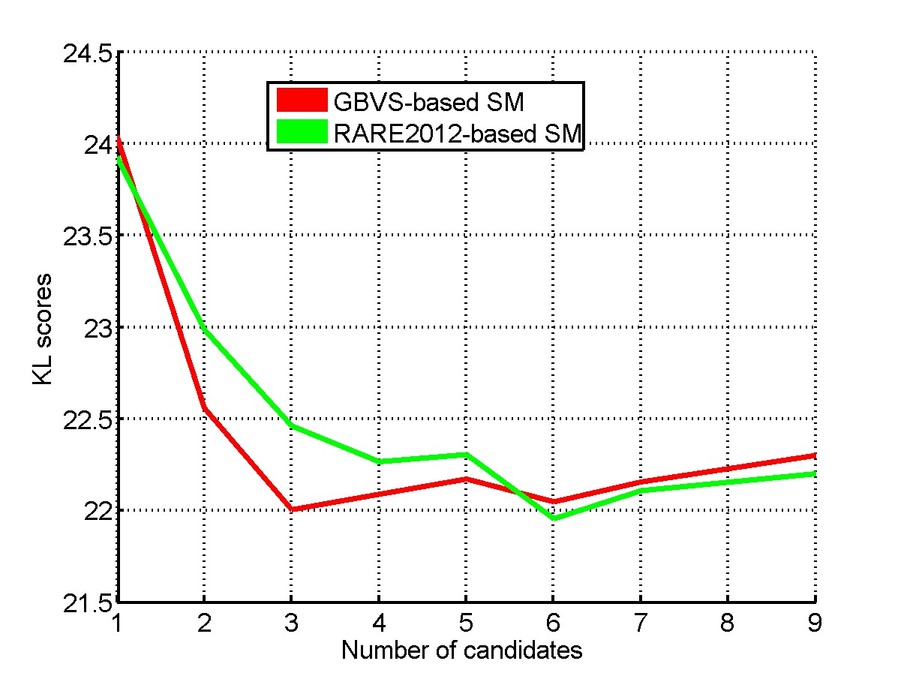}}    
    \subfigure[4-6 years-old]{\includegraphics[width=0.325\linewidth]{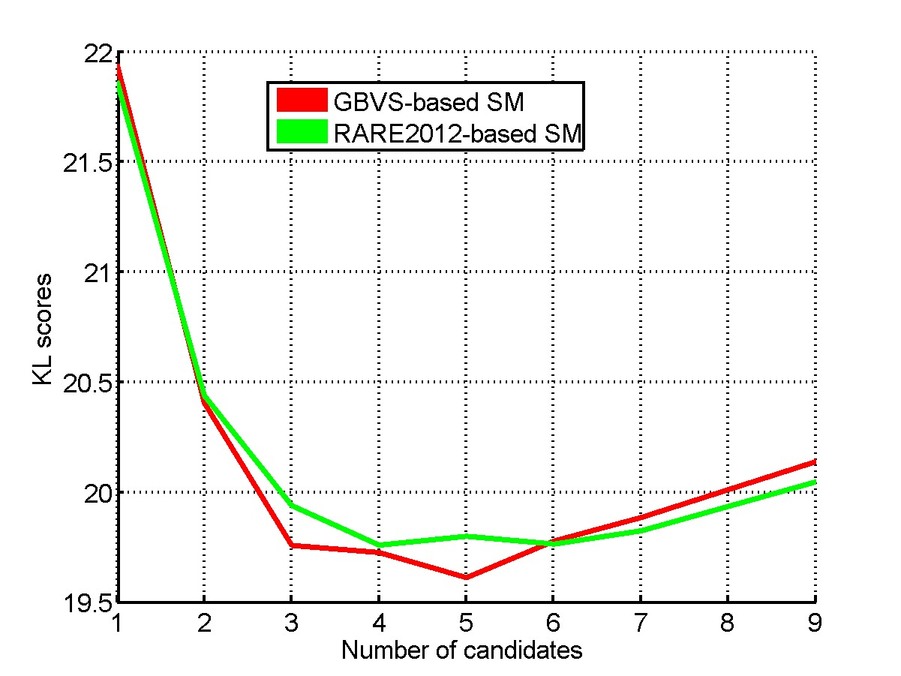}}
    \subfigure[Adults]{\includegraphics[width=0.325\linewidth]{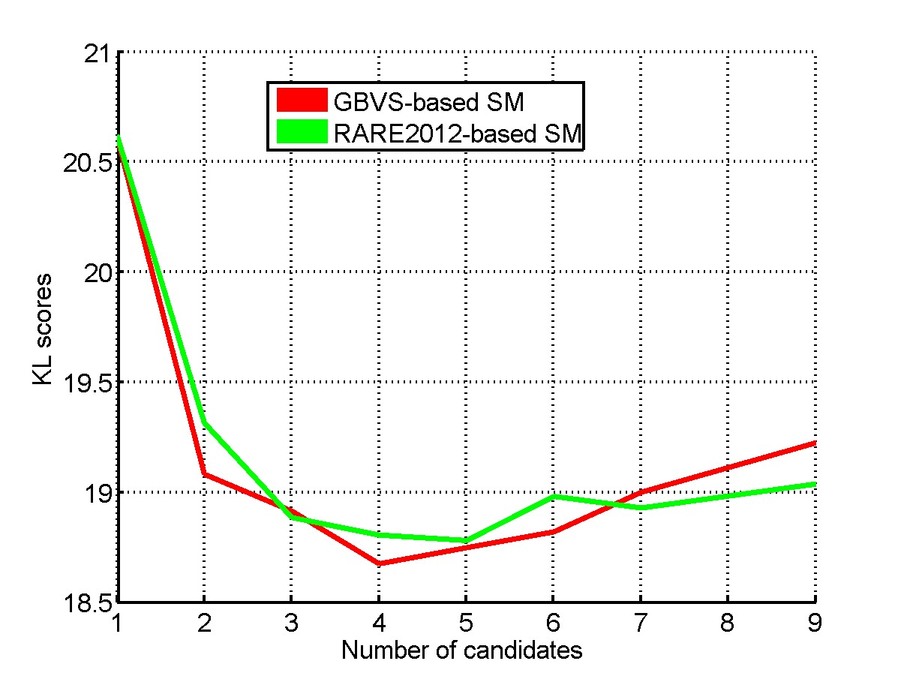}}
  \captionof{figure}{KL-divergence between the actual and the predicted distributions in function of the number of candidates $N_c$ for the adults, 4-6 years-old and 2 years-old groups. Top row: the KL-divergence is computed between the actual distribution of saccade ampltiudes and the predicted one. Bottom-row: the KL-divergence is computed between the actual joint distribution of saccade amplitudes and saccade orientations and the predicted one.}\label{fig:predictedScanpathsDistributionKL}
\end{figure*}

Saccadic models aim to predict salient areas as well as to generate scanpaths that present similar features as human scanpaths. From the predicted scanpaths, we compute, for each age group and for both saliency models (i.e. GBVS and RARE2012), the 1D distribution of saccade amplitudes and the 2D joint distribution of saccade amplitudes and saccade orientations. We evaluate the Kullback-Leibler (KL) divergence between these distributions and the distributions computed from eye tracking data. \figurename~\ref{fig:predictedScanpathsDistributionKL} plots the KL scores in function of the parameter $N_c$. We observe that the KL scores follow a U-shaped curve. The KL scores are higher for low and high values of $N_c$. A low value of $N_c$ corresponds to high dispersion between observers whereas a high value reduces the randomness of the fixation point selection.  The best KL scores are obtained for $N_c$ in the range 4 to 6. 
More specifically, for each age group, we select the best $N_c$ value in order to get the best compromise between salient area prediction and scanpath plausibility.  For adults and 6-10 year-old groups, $N_c=4$. For 4-6 and 2 year-old, $N_c=5$. 

\begin{figure*}
\centering  
	\subfigure{\includegraphics[width=0.325\linewidth]{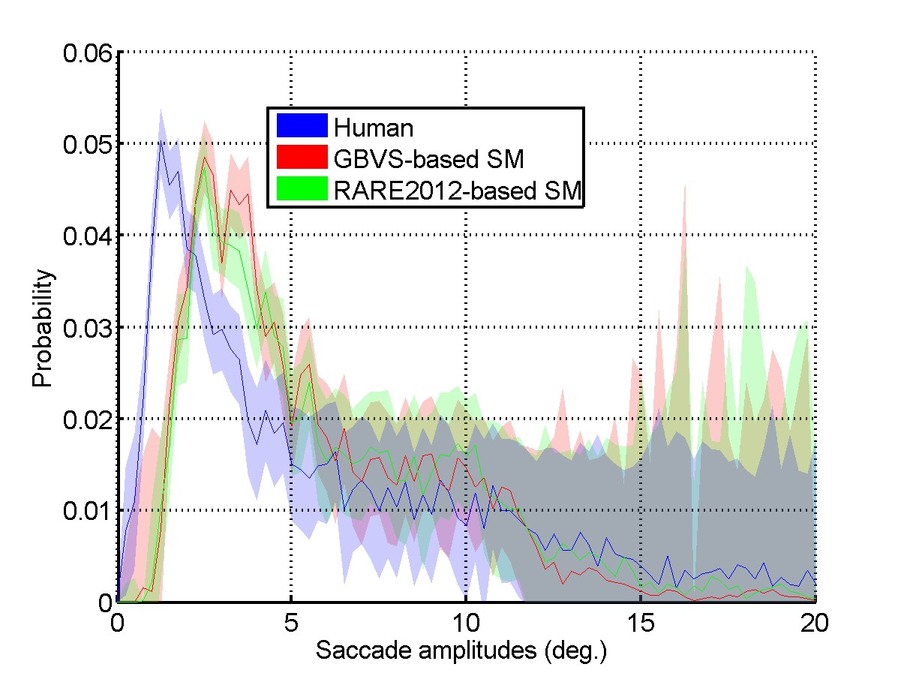}}    
	\subfigure{\includegraphics[width=0.325\linewidth]{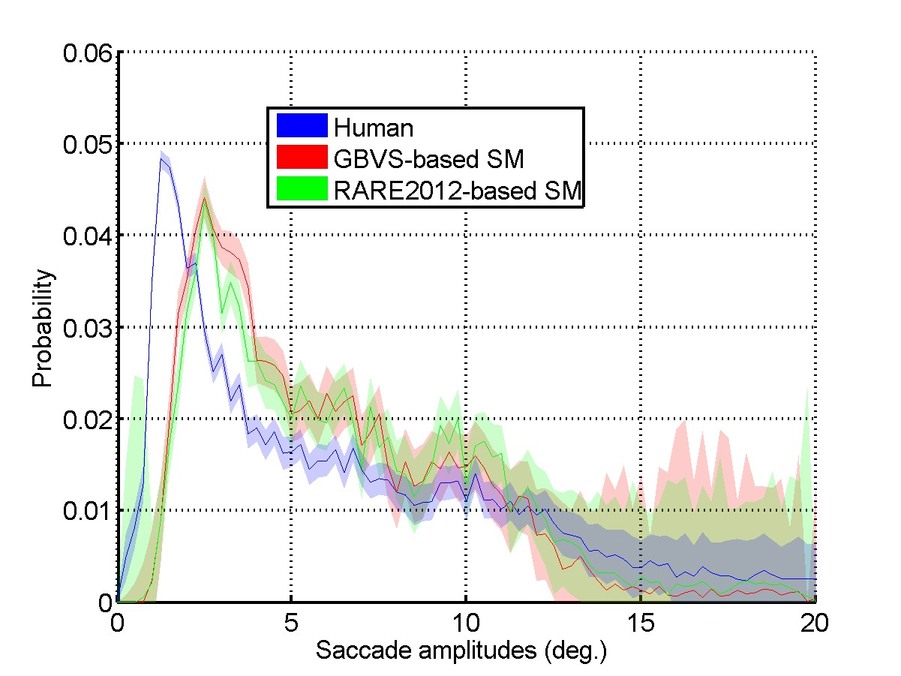}}        
    \subfigure{\includegraphics[width=0.325\linewidth]{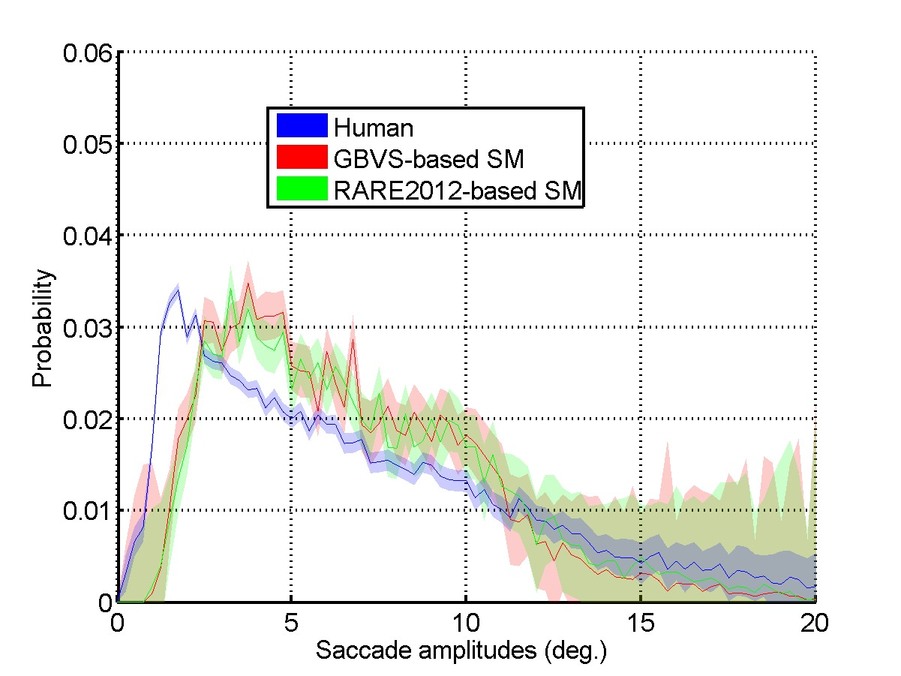}}         
    \subfigure{\includegraphics[width=0.325\linewidth]{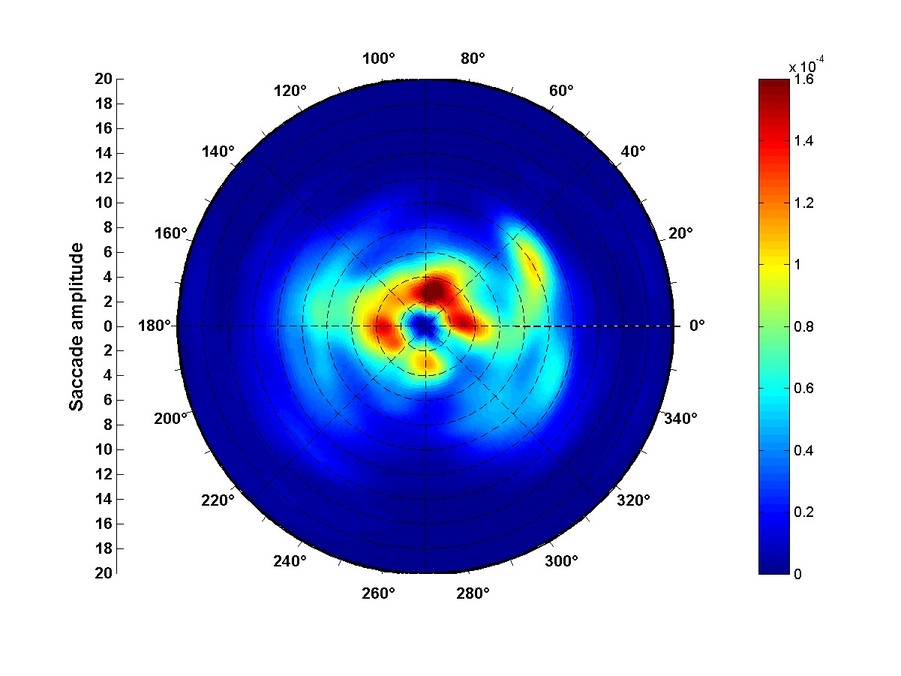}}    
    \subfigure{\includegraphics[width=0.325\linewidth]{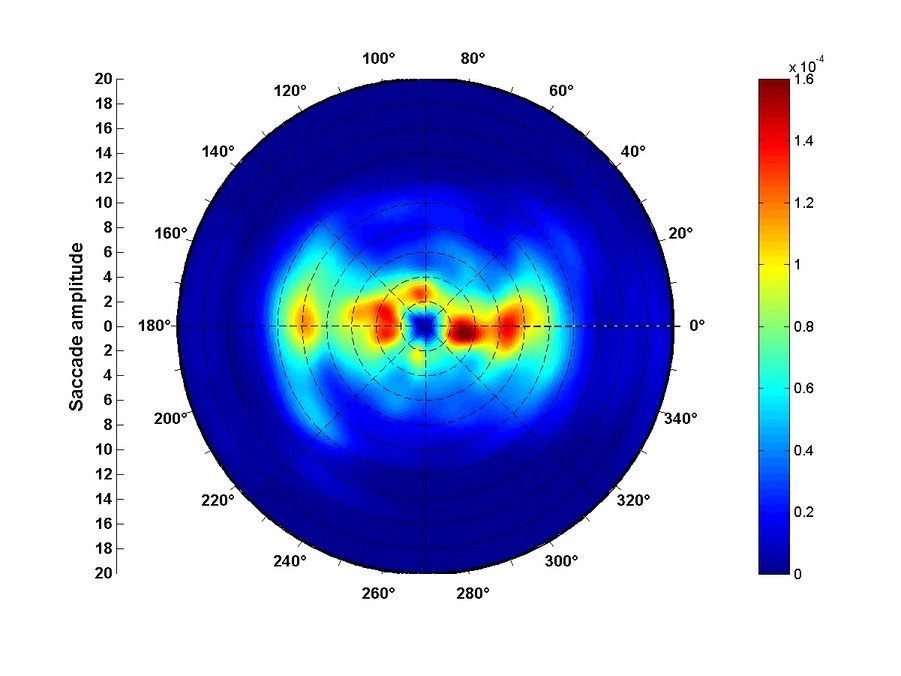}}    
    \subfigure{\includegraphics[width=0.325\linewidth]{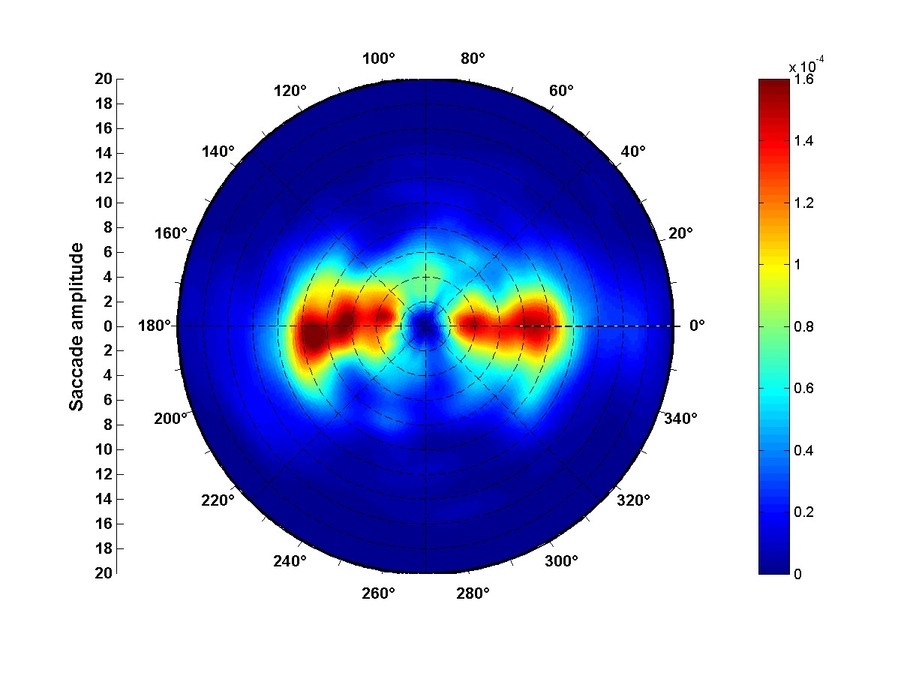}}
    \setcounter{subfigure}{0}
    \subfigure[2 years-old]{\includegraphics[width=0.325\linewidth]{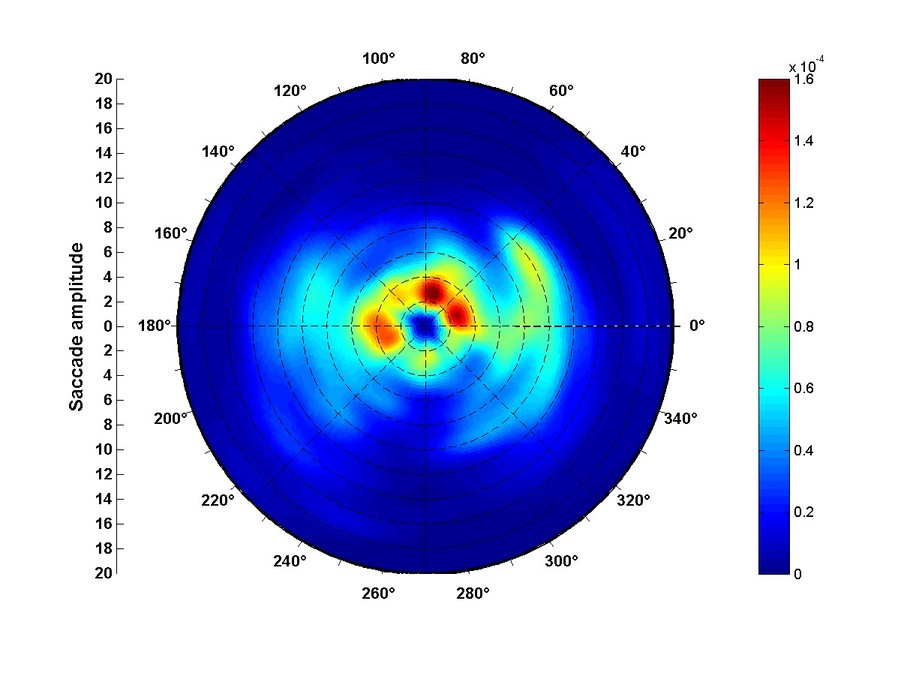}}    
    \subfigure[4-6 years-old]{\includegraphics[width=0.325\linewidth]{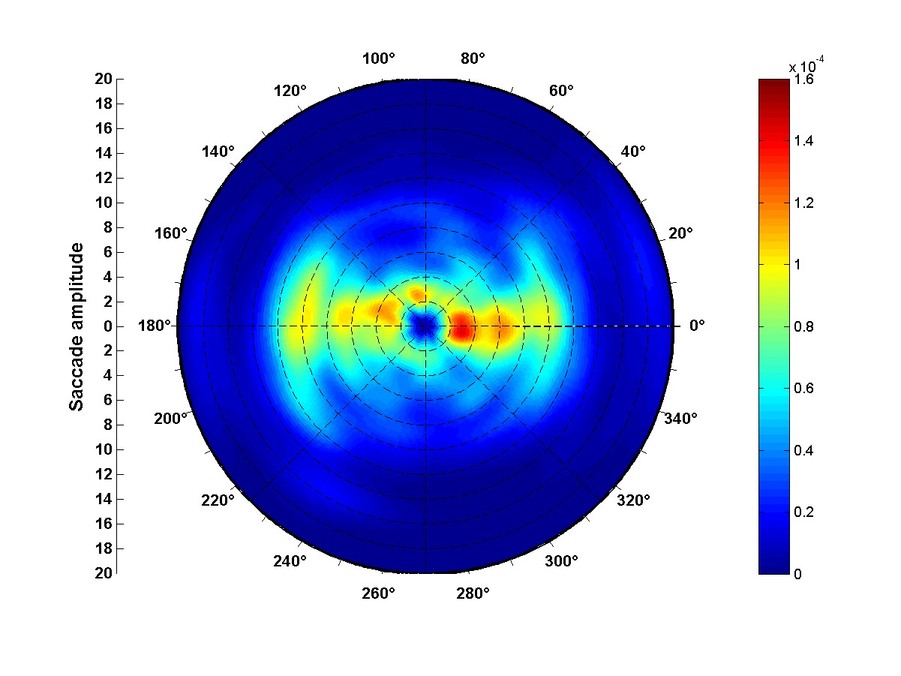}}    
    \subfigure[Adults]{\includegraphics[width=0.325\linewidth]{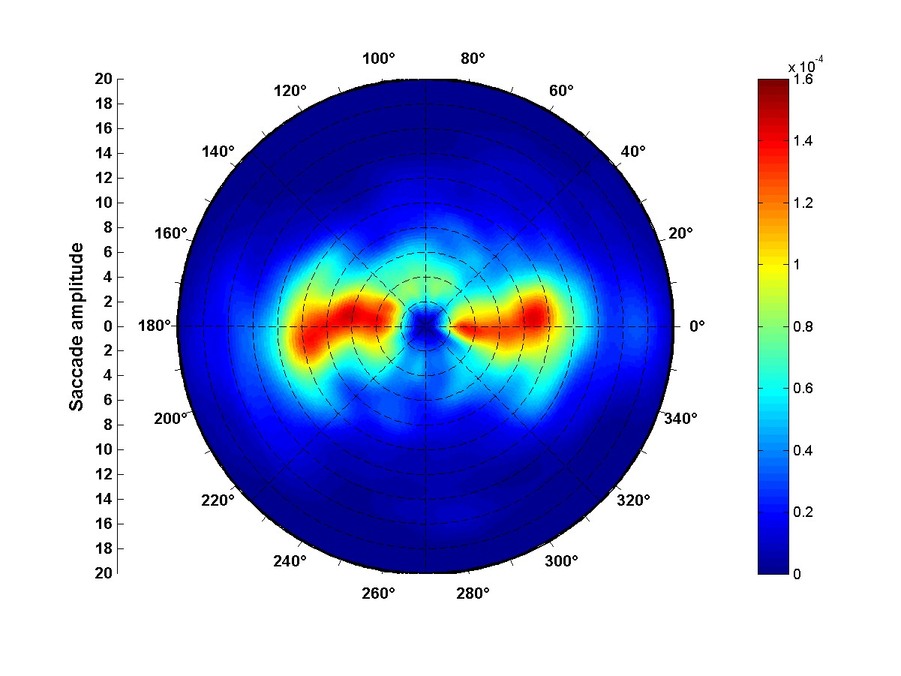}}
  \captionof{figure}{Features of the predicted scanpaths. Top row: the actual and the predicted distributions of saccade amplitudes are plotted for $N_c=5$, $N_c=4$ and $N_c=4$ corresponding to 2 year-old, 4-6 year-old and adults groups, respectively. Middle and bottom rows: joint distributions of saccade amplitudes and saccade orientations computed from GBVS-based saccadic model and RARE2012-based saccadic model, respectively.}\label{fig:predictedDistribution}
\end{figure*}

\figurename~\ref{fig:predictedDistribution} shows the distributions of saccade amplitudes (top row) and the joint distributions of saccade amplitudes and orientations for the age groups when considering the aforementioned values of $N_c$ (bottom row)\footnote{In supplementary material, more results are given, especially for low and high values of $N_c$.}. We observe that the distributions of saccade amplitudes computed with the proposed saccadic model have a similar shape when compared with actual distributions. We note, however, that the proposed model tends to generate larger saccades. The main peak of the predicted distributions is between 2 and 3 degrees of visual angle, whereas the main peak of actual distributions is about 2 degrees of visual angle. This discrepancy might be due to the computational modelling of the inhibition-of-return mechanism which does not entirely reflect the reality. A second explanation might be related to the computation of joint distributions, as well as how they are used. One of the strength of the proposed saccadic model  is that we use  spatially-variant joint distributions (see section~\ref{sec:discussion} for more details). However, only 9 joint distributions are used to reproduce the gaze deployment, which might not be enough. Increasing this number would make sense but would require more fixation points in order to compute accurate and relevant distributions. Another concern pertains to the memory effect that is not taken into account. Indeed, there is a time dependency in saccade amplitudes. Small amplitude saccades tend to be followed by large amplitude saccades, which are followed by small ones~\cite{Unema2005,Tatler2008}. The plots in~\figurename~\ref{fig:predictedDistribution} also show that the key ingredient to produce plausible scanpath is not the input saliency map. Although that GBVS and RARE2012 models generate saliency maps that have rather different saliency distributions, as illustrated by~\figurename~\ref{fig:predictedSaliencyMap},  the saccadic model manages to produce plausible scanpaths in both cases.

The middle and bottom rows of~\figurename~\ref{fig:predictedDistribution} illustrate the joint distributions computed from GBVS-based scanpaths and RARE2012-based scanpaths, respectively. Compared to actual joint distributions shown in~\figurename~\ref{fig:etJointDistribution}, we observe a similar evolution of the saccadic behavior. For the 2 year-old group, saccade amplitudes are rather small and isotropic. The horizontal bias as well as large saccades progressively appears with aging. The horizontal bias is very noticeable for adults groups. 
 
\subsection{Joint distribution influences}\label{sec:discussion}
In this section, we discuss the influence of the joint distributions by comparing the performance of the proposed age-dependent saccadic model with those obtained by considering age-independent distribution, uniform joint distribution and spatially-invariant distribution. We perform these tests by considering the following setting: GBSV and RARE2012 model, adult groups and $N_c=4$.

\subsubsection{Age-dependent vs age-independent distribution}
In this case, instead of using the spatially-variant joint distribution of adult group, we use the 2 y.o. spatially-variant joint distribution when computing adult scanpaths. We evaluate the performance of this modified model with the adult ground truth. Table~\ref{table:perfSaliencyModels} (bottom row called SVDist2yo) indicates that the ability to predict salient areas decreases when considering 2 y.o. distribution instead of adult one.  In addition, when comparing the saccade amplitude distribution generated by this model (see~\figurename~\ref{fig:nonPlausibleJointDistribution}) with the best one (see top-right plot in~\figurename~\ref{fig:predictedDistribution}), we observe that the predicted scanpaths are less plausible than those obtained with the model using adult distribution.  

\subsubsection{Uniform joint distribution}
\begin{figure*}
\setcounter{subfigure}{0}
\centering  
    \subfigure[]{\includegraphics[width=0.45\linewidth]{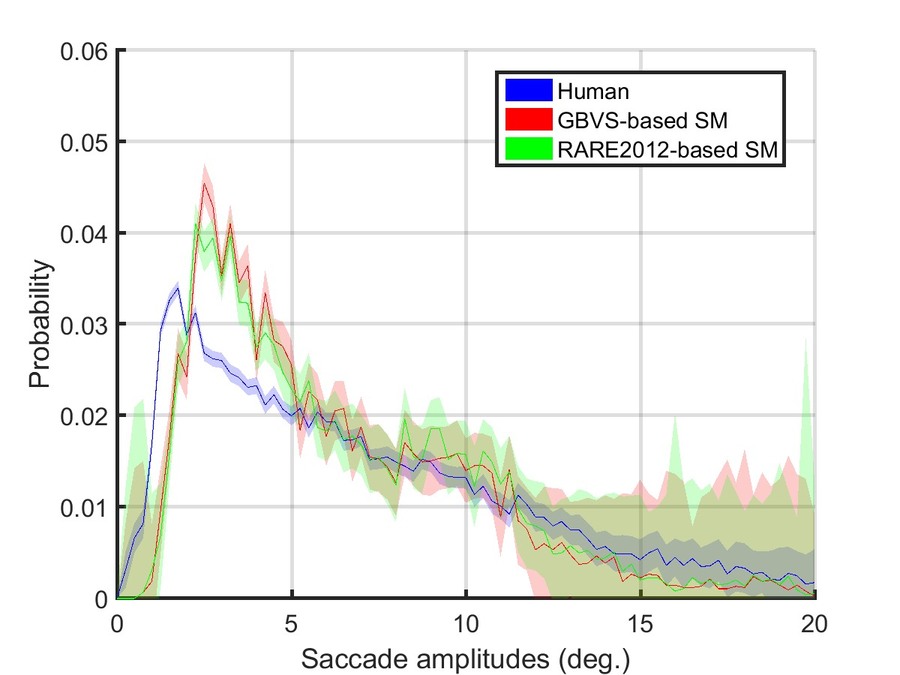}}
	\subfigure[]{\includegraphics[width=0.45\linewidth]{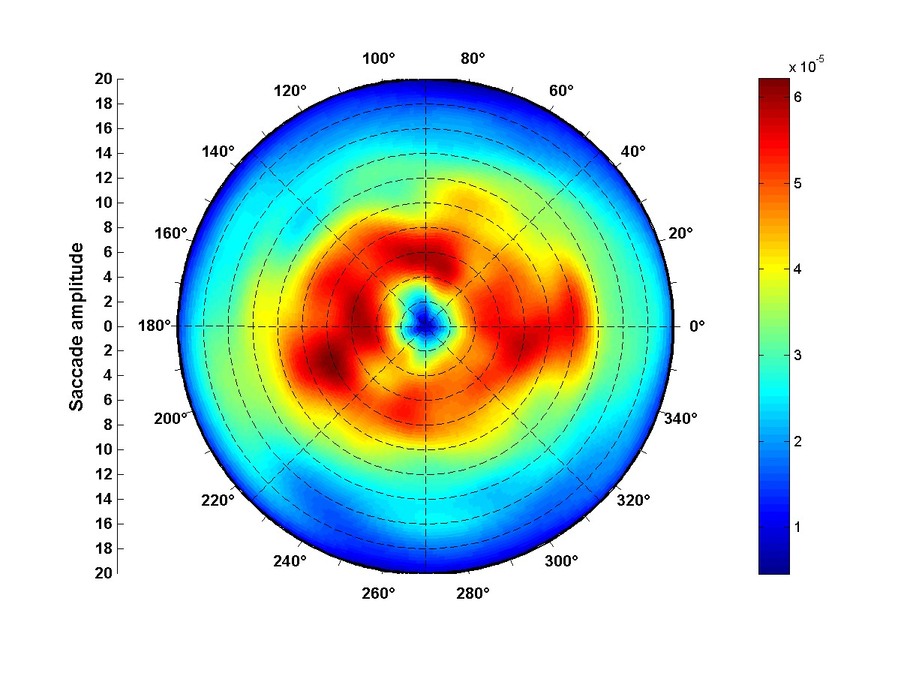}}    
  \captionof{figure}{(a) Actual and the predicted distributions of saccade amplitudes (see the difference with the top-right distribution in~\figurename~\ref{fig:predictedDistribution});(b) Joint distribution of saccade amplitudes and orientations when we do not consider viewing biases: $p_B(d(x,x_{t-1}),\phi(x,x_{t-1}))=1$, $\forall x\in\Omega$. Note that the scale is not similar to previous used scales.}\label{fig:nonPlausibleJointDistribution}
\end{figure*}

To further evaluate the influence of the joint distribution on the results, we set in equation~\ref{equ:model}, $p_B(d(x,x_{t-1}),\phi(x,x_{t-1}))=1$, $\forall x\in\Omega$. In~\cite{Tatler2009}, Tatler and Vincent gave evidence that the viewing biases may be fundamental to predict where we look at. Results are presented in the bottom of Table~\ref{table:perfSaliencyModels}. As expected, the performances decrease, but they are still interesting. However, this solution does not allow us to generate plausible visual scanpaths as illustrated in~\figurename~\ref{fig:nonPlausibleJointDistribution}. 

\subsubsection{Spatially-variant vs invariant joint distributions}
As presented at the bottom of Table~\ref{table:perfSaliencyModels}, the use of spatially-variant joint distributions increases  the performance of the saccadic model when compared to the saccadic model using a non spatially-variant joint distribution (see the acronym \textit{NSV dist.}).

%Training effect see~\cite{LeMeur2015} does not harm the performance since the joint distribution of saccade amplitudes and orientations represents the systematic tendencies that are common across a group of observers.

\subsection{Generated saliency map}
\figurename~\ref{fig:SaliencyMap} presents the human saliency maps for the 2 year-old (b) and adult group (c). The saliency maps predicted by \textit{traditional} saliency models are also illustrated in (d) and (e), corresponding to the GBVS~\cite{Harel2006} and RARE2012 model~\cite{Riche2013}, respectively.

When comparing 2 year-old and adult saliency maps, we observe two main differences. First, adults explore much more the visual scene than 2 year-old children. Saliency maps of adults are therefore less focused than those of children. The second observation concerns the center bias which is more pronounced for 2 year-old group (see also~\figurename~\ref{fig:centerBias} for a quantitative measure).

\subsection{Influence of the parameter $N_c$}\label{sec:nc}
\subsubsection{On saliency map}
\begin{figure*}[h]
\centering 
    \subfigure{\includegraphics[width=0.19\linewidth]{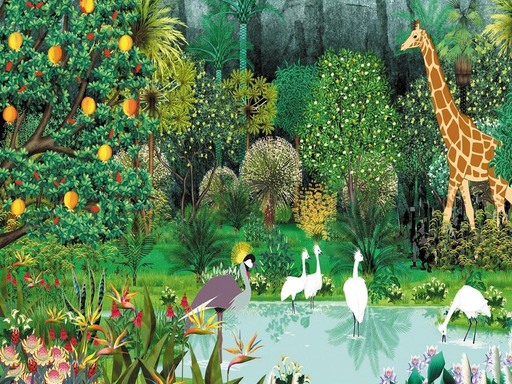}}
    \subfigure{\includegraphics[width=0.19\linewidth]{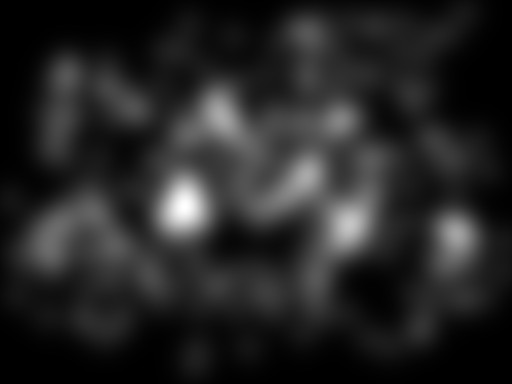}}
    \subfigure{\includegraphics[width=0.19\linewidth]{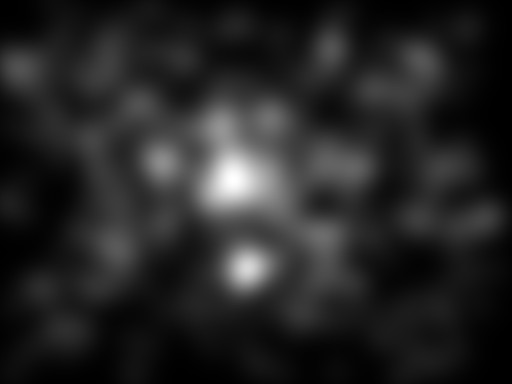}}
    \subfigure{\includegraphics[width=0.19\linewidth]{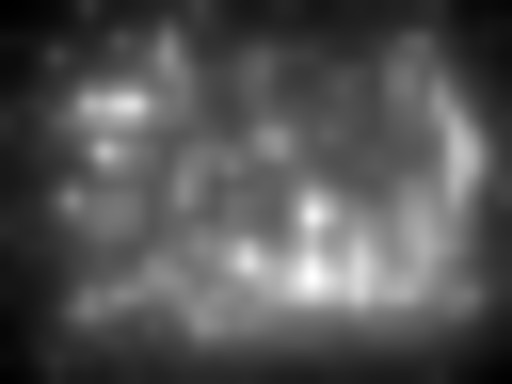}}    
    \subfigure{\includegraphics[width=0.19\linewidth]{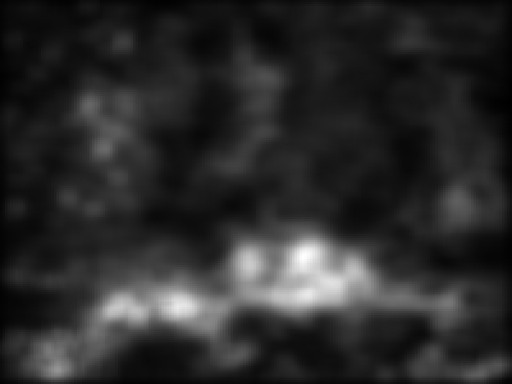}}
    
    \subfigure{\includegraphics[width=0.19\linewidth]{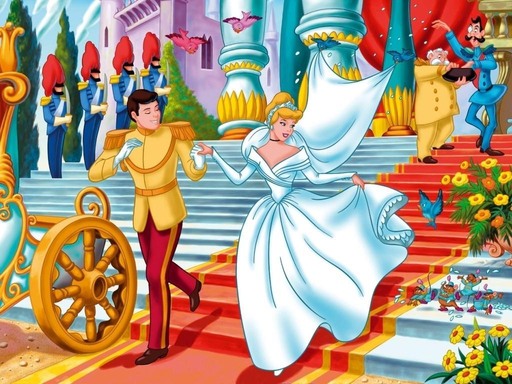}}
    \subfigure{\includegraphics[width=0.19\linewidth]{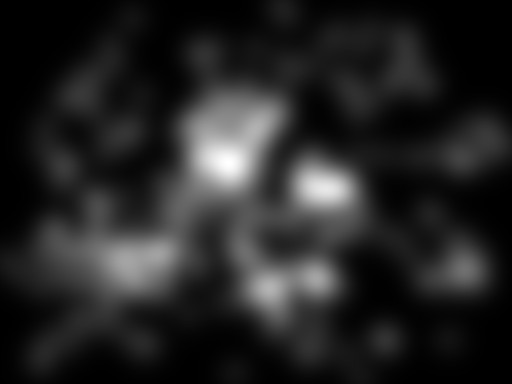}}
    \subfigure{\includegraphics[width=0.19\linewidth]{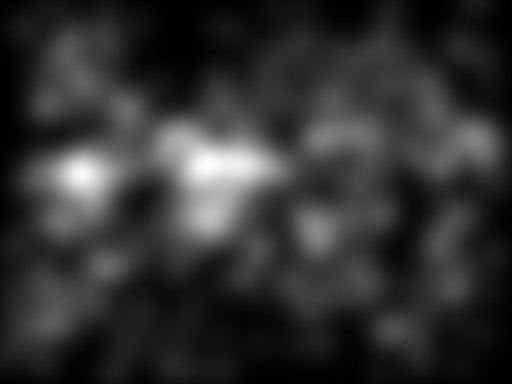}}
    \subfigure{\includegraphics[width=0.19\linewidth]{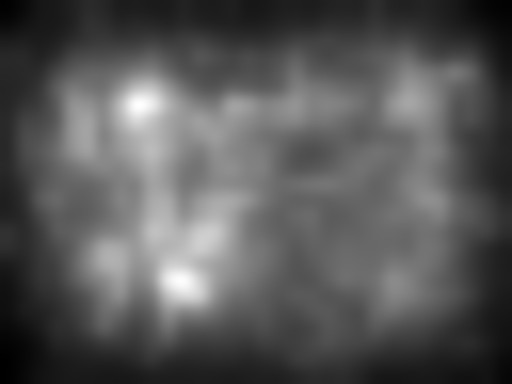}}    
    \subfigure{\includegraphics[width=0.19\linewidth]{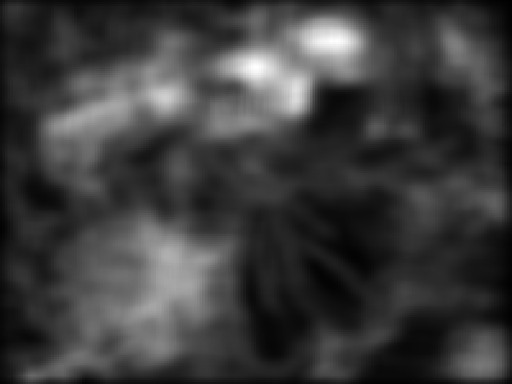}}
    
    \subfigure{\includegraphics[width=0.19\linewidth]{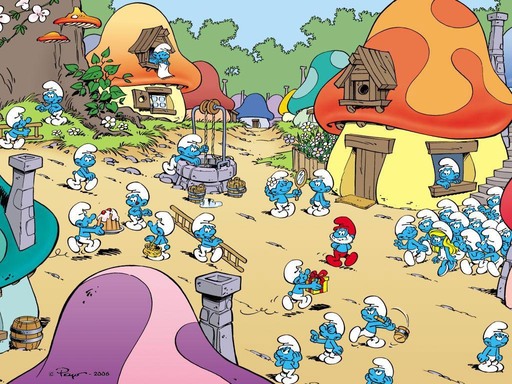}}
    \subfigure{\includegraphics[width=0.19\linewidth]{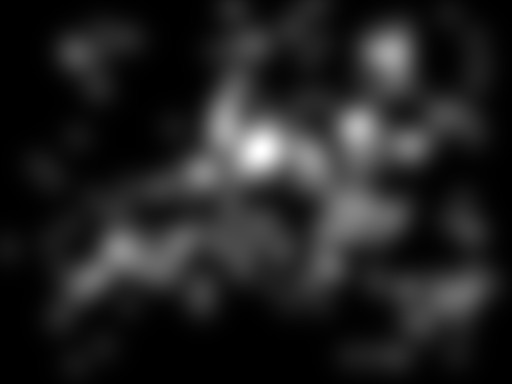}}
    \subfigure{\includegraphics[width=0.19\linewidth]{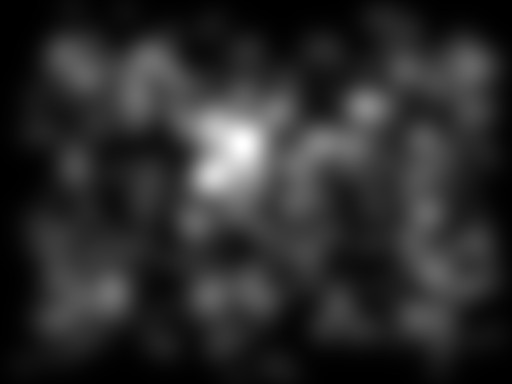}}
    \subfigure{\includegraphics[width=0.19\linewidth]{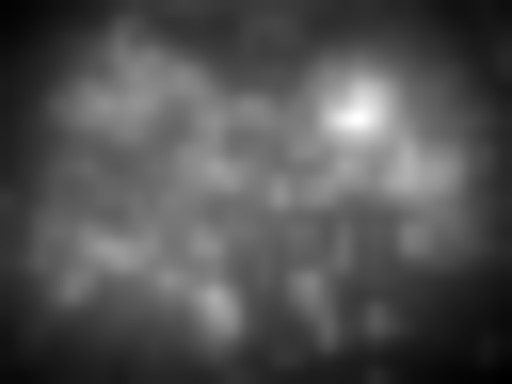}}    
    \subfigure{\includegraphics[width=0.19\linewidth]{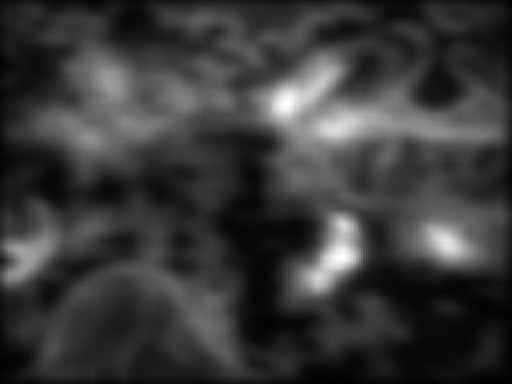}}
    \setcounter{subfigure}{0}
    \subfigure[Orig.]{\includegraphics[width=0.19\linewidth]{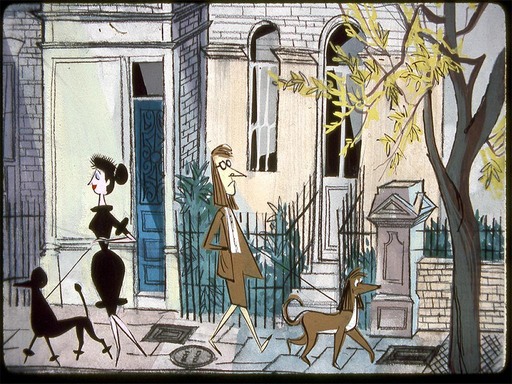}}
    \subfigure[Human 2 yo]{\includegraphics[width=0.19\linewidth]{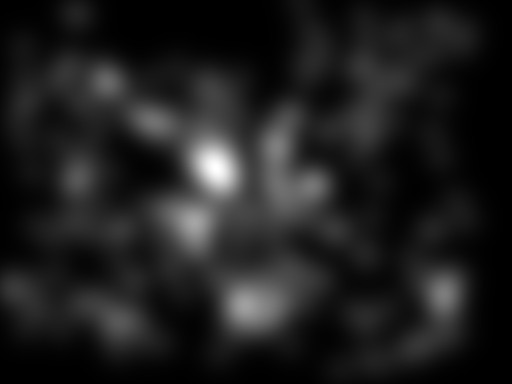}}
    \subfigure[Human adult] {\includegraphics[width=0.19\linewidth]{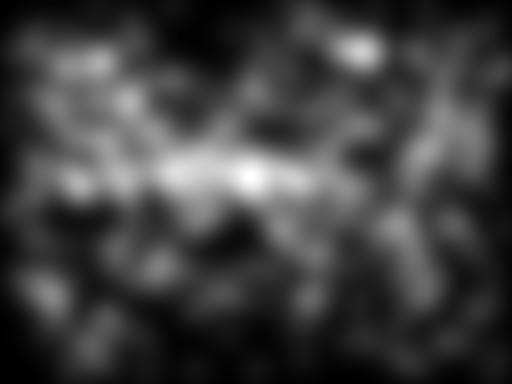}}
    \subfigure[GBVS]{\includegraphics[width=0.19\linewidth]{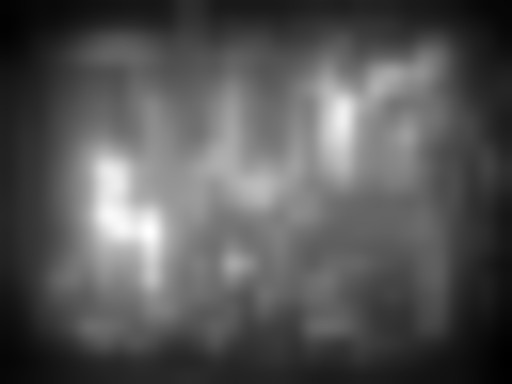}}    
    \subfigure[RARE2012]{\includegraphics[width=0.19\linewidth]{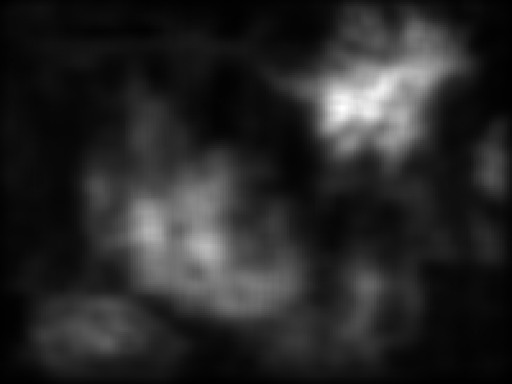}}       
  \captionof{figure}{(a) Original image; (b) and (c) represent the human saliency maps for 2 year old and adult groups, respectively; (d) and (e) represent the saliency mpas predicted by GBVS~\cite{Harel2006} and RARE2012~\cite{Riche2013} saliency model, respectively.}\label{fig:SaliencyMap}
\end{figure*}

\begin{figure*}[h]
\centering 
   
    \subfigure{\includegraphics[width=0.135\linewidth]{supp/sm/P1}}
    \subfigure{\includegraphics[width=0.135\linewidth]{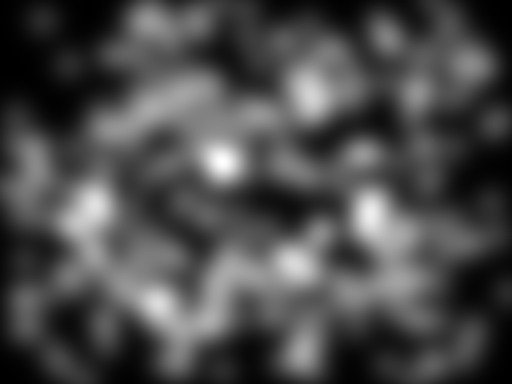}}
    \subfigure{\includegraphics[width=0.135\linewidth]{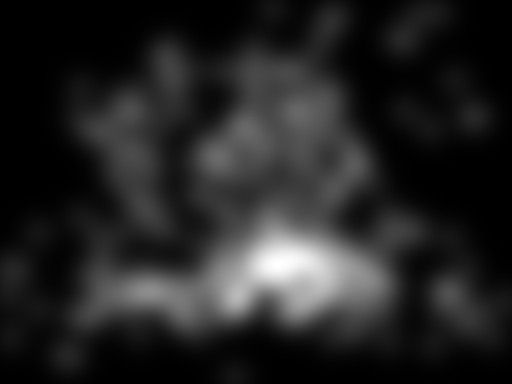}}
    \subfigure{\includegraphics[width=0.135\linewidth]{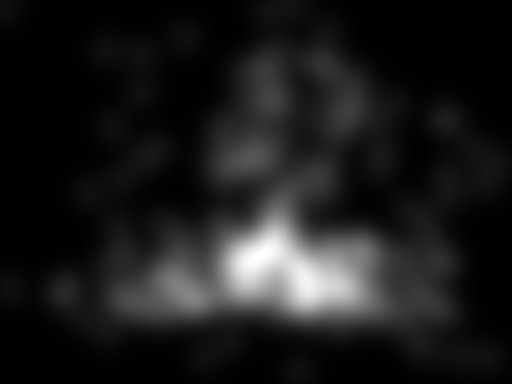}}    
    \subfigure{\includegraphics[width=0.135\linewidth]{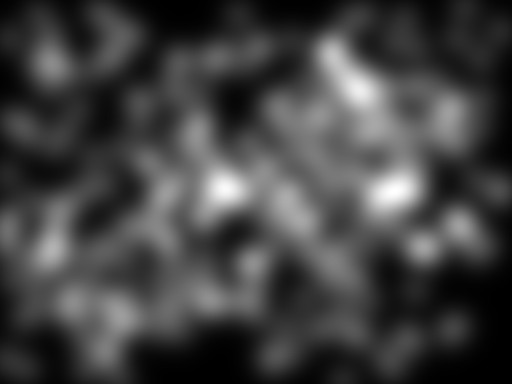}}
    \subfigure{\includegraphics[width=0.135\linewidth]{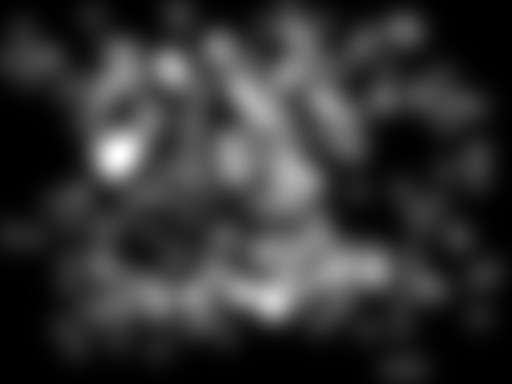}}
    \subfigure{\includegraphics[width=0.135\linewidth]{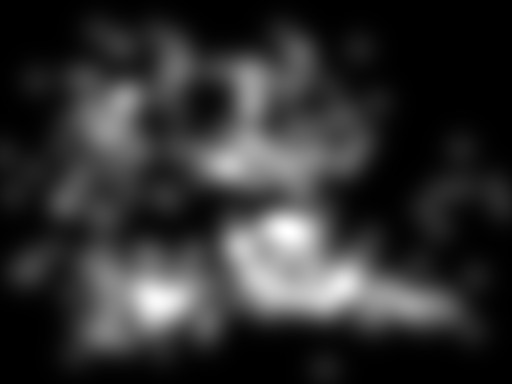}}
    
    \subfigure{\includegraphics[width=0.135\linewidth]{supp/sm/P3}}
    \subfigure{\includegraphics[width=0.135\linewidth]{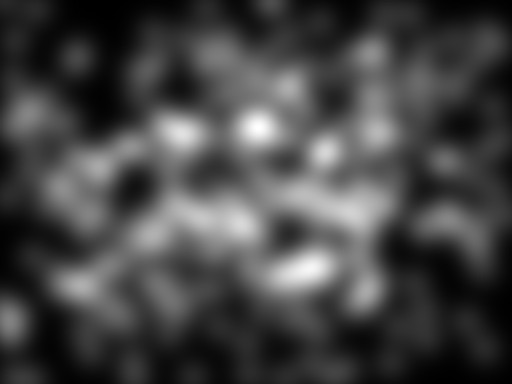}}
    \subfigure{\includegraphics[width=0.135\linewidth]{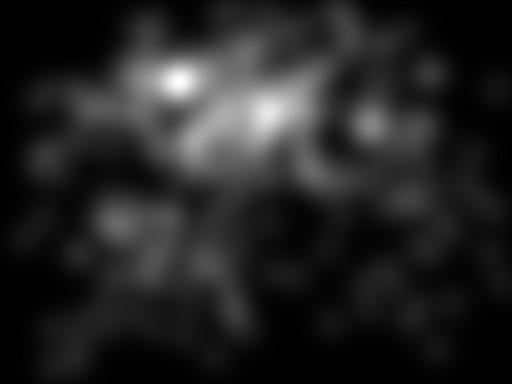}}
    \subfigure{\includegraphics[width=0.135\linewidth]{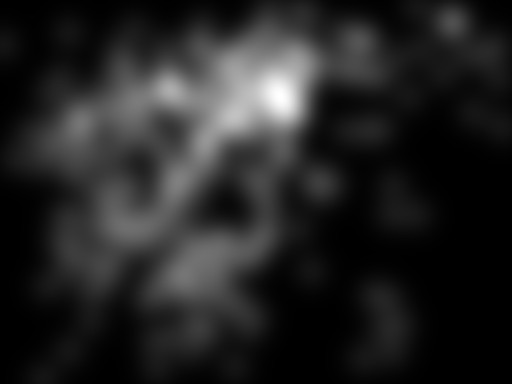}}    
    \subfigure{\includegraphics[width=0.135\linewidth]{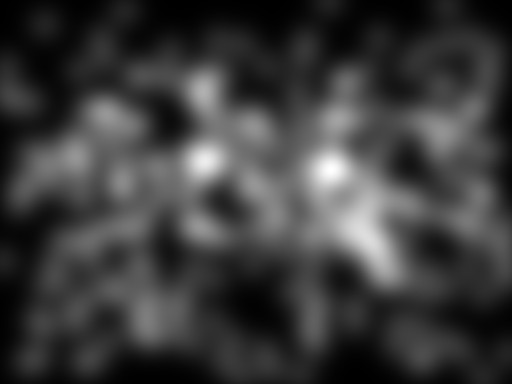}}
    \subfigure{\includegraphics[width=0.135\linewidth]{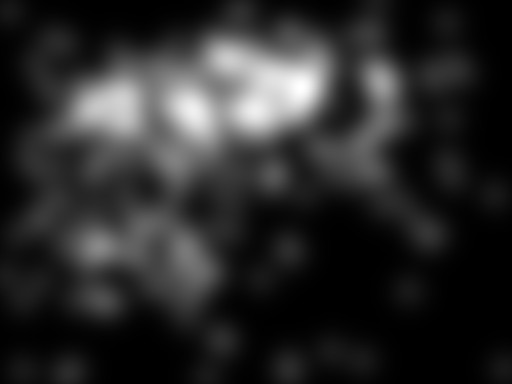}}
    \subfigure{\includegraphics[width=0.135\linewidth]{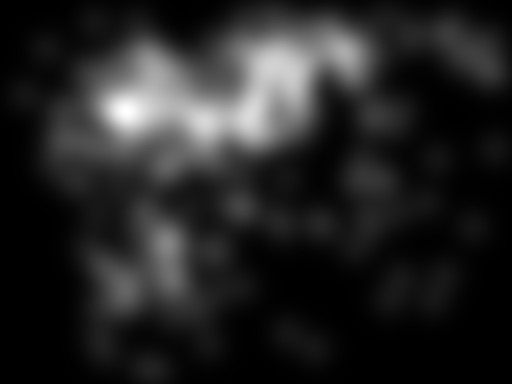}}

	\subfigure{\includegraphics[width=0.135\linewidth]{supp/sm/P17}}
    \subfigure{\includegraphics[width=0.135\linewidth]{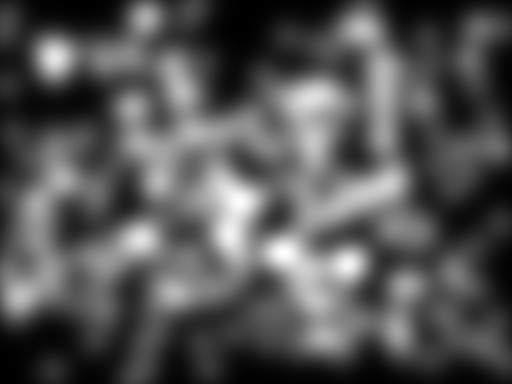}}
    \subfigure{\includegraphics[width=0.135\linewidth]{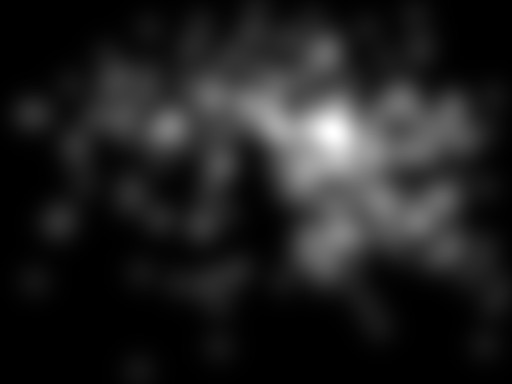}}
    \subfigure{\includegraphics[width=0.135\linewidth]{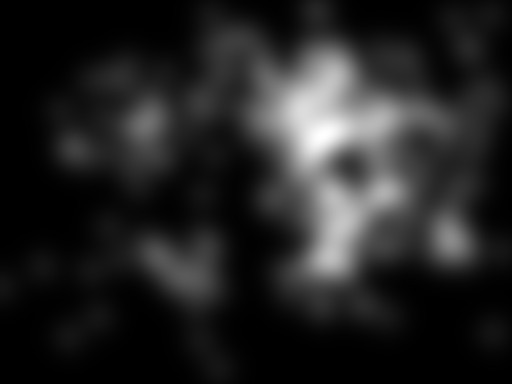}}    
    \subfigure{\includegraphics[width=0.135\linewidth]{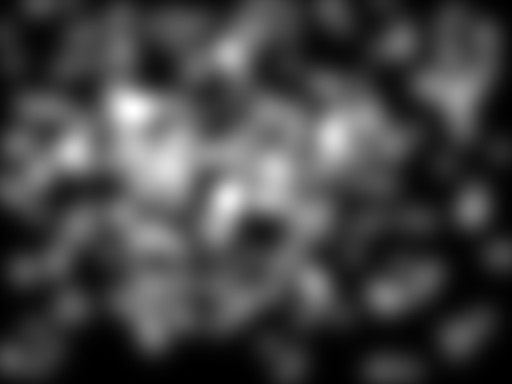}}
    \subfigure{\includegraphics[width=0.135\linewidth]{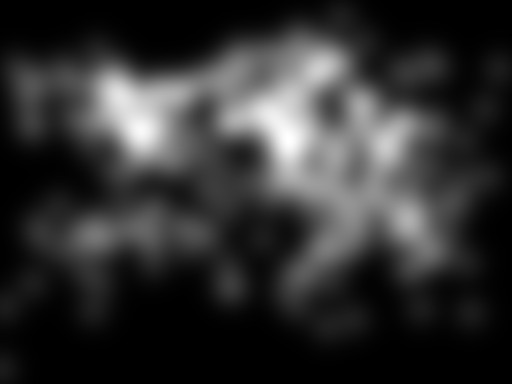}}
    \subfigure{\includegraphics[width=0.135\linewidth]{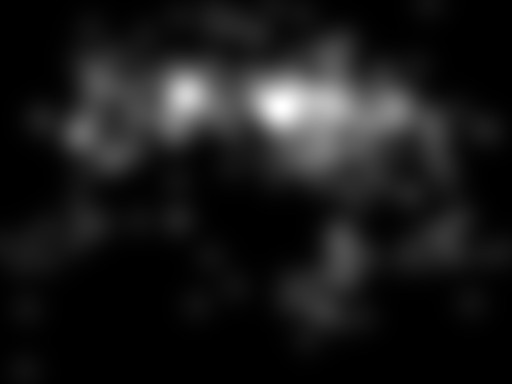}}
    
    \setcounter{subfigure}{0}
    \subfigure[Orig.]{\includegraphics[width=0.135\linewidth]{supp/sm/P4}}
    \subfigure[$N_c=1$, 2 yo]{\includegraphics[width=0.135\linewidth]{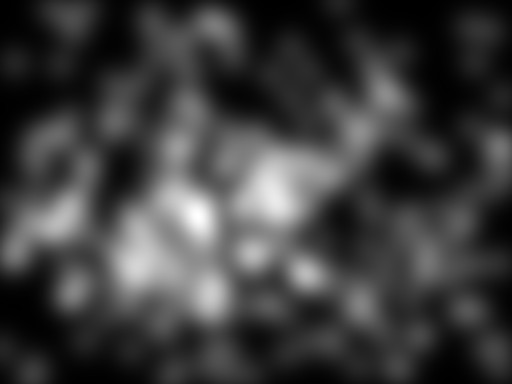}}
    \subfigure[Best 2 yo]{\includegraphics[width=0.135\linewidth]{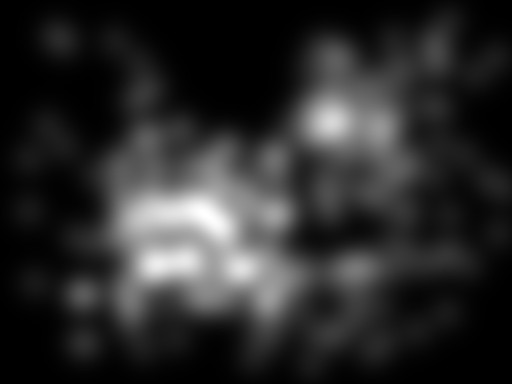}}
    \subfigure[$N_c=9$, 2 yo]{\includegraphics[width=0.135\linewidth]{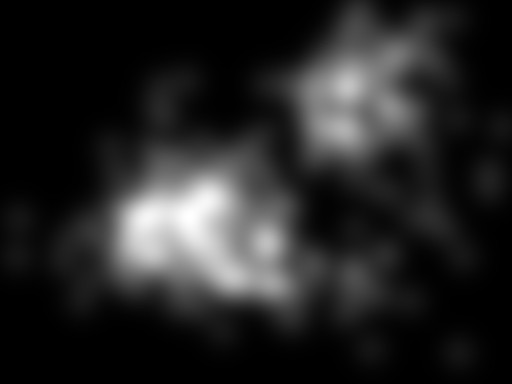}}    
    \subfigure[$N_c=1$, Adults]{\includegraphics[width=0.135\linewidth]{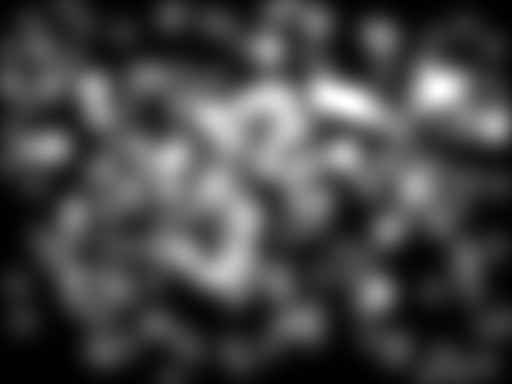}}
    \subfigure[Best adults]{\includegraphics[width=0.135\linewidth]{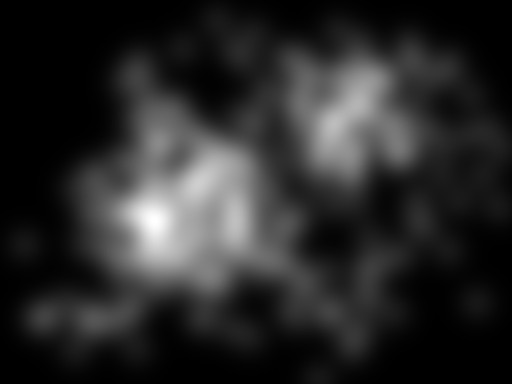}}
    \subfigure[$N_c=9$, Adults]{\includegraphics[width=0.135\linewidth]{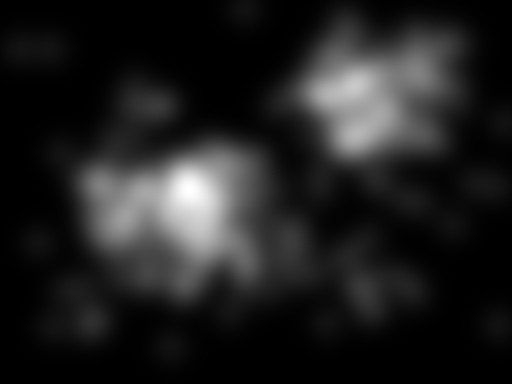}}
    
  \captionof{figure}{Influence of the parameter $N_c$ on the scanpath-based saliency maps. (a) Original image; (b) to (d) illustrate the saliency map for the 2 year-old group for $N_c=1$, the best value of $N_c$ (5) and $N_c=9$, respectively; (e) to (g) illustrate the saliency map for adult group for $N_c=1$, the best value of $N_c$ (4) and $N_c=9$, respectively.}\label{fig:predictedSaliencyMapSP}
\end{figure*}

\figurename~\ref{fig:predictedSaliencyMapSP} illustrates the influence of the parameter $N_c$ on the scanpath-based saliency maps. These maps are obtained when the input saliency maps are computed by RARE2012 model. We observe that, when $N_c=1$, the salience dispersion is high. At the opposite, increasing the value of $N_c$ reduces the dispersion and provides more focused saliency maps. The best results are obtained for $N_c=5$ and $N_c=4$ for 2 year-olds and adults, respectively. It comes out that predicted saliency maps for adults group are less focused than predicted saliency maps for 2 year-old group, which is in agreement with the previous observations (see \figurename~\ref{fig:SaliencyMap} (b) and (c)).

\subsubsection{On predicted scanpaths}
\begin{figure*}[h]
\centering 
   
    \subfigure{\includegraphics[width=0.32\linewidth]{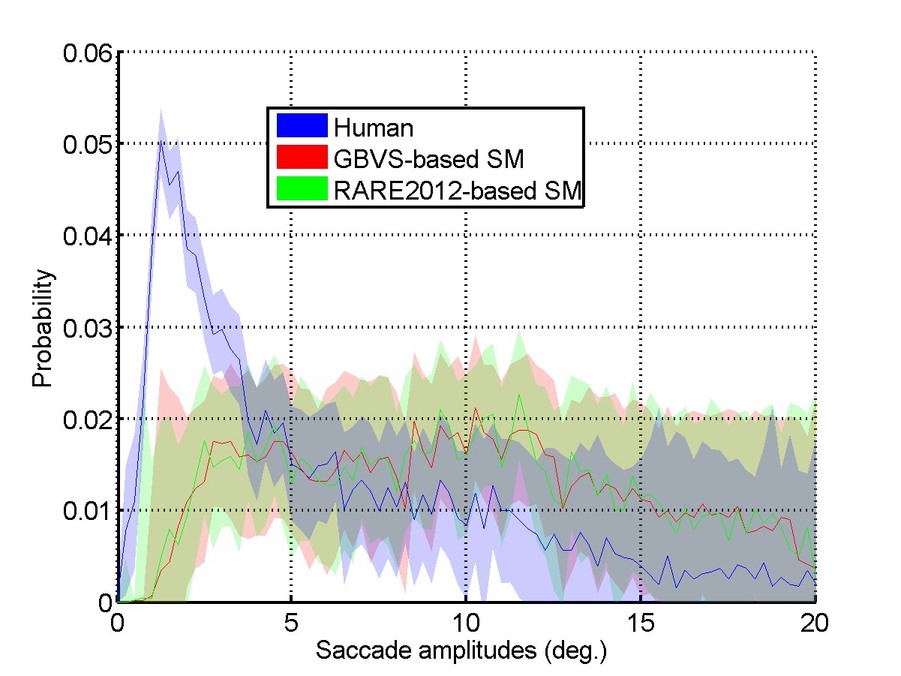}}
    \subfigure{\includegraphics[width=0.32\linewidth]{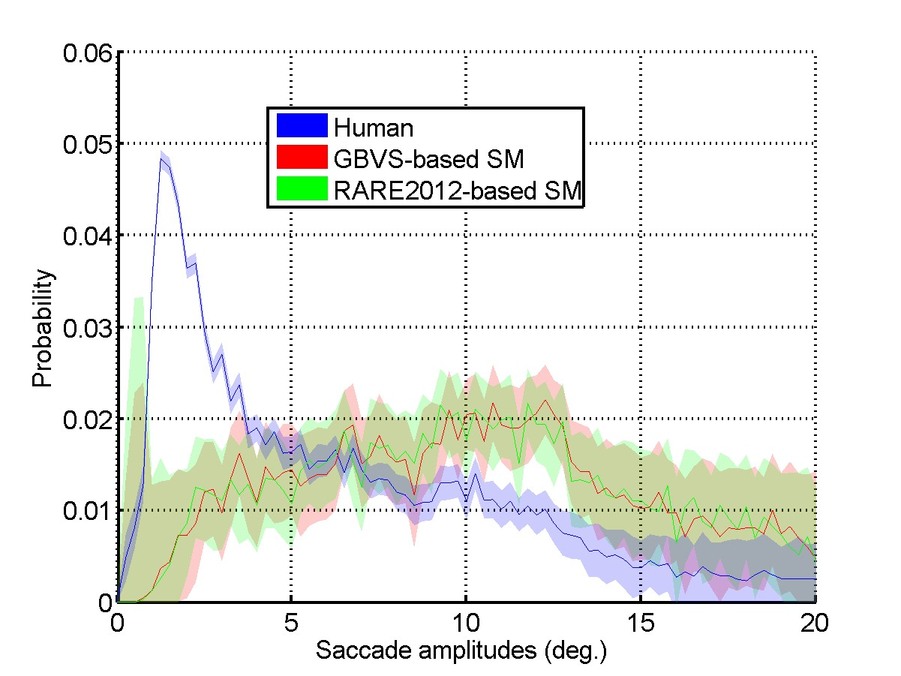}}
    \subfigure{\includegraphics[width=0.32\linewidth]{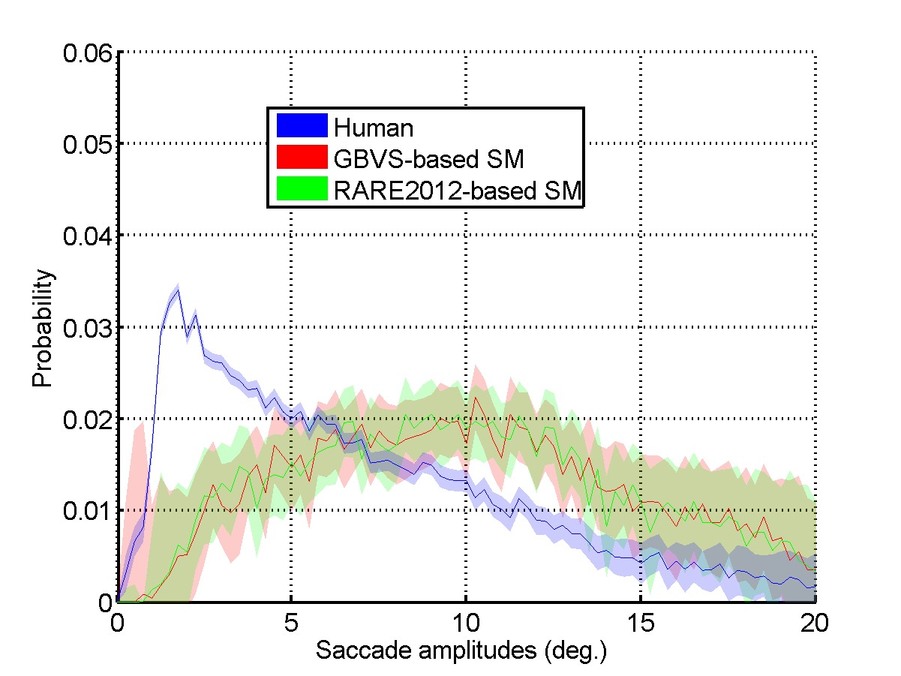}}
      
    \subfigure{\includegraphics[width=0.32\linewidth]{prediction/simuNc5_2yo_amplitudeSaccade_A1}}  
    \subfigure{\includegraphics[width=0.32\linewidth]{prediction/simuNc5_46yo_amplitudeSaccade_A1}}  
    \subfigure{\includegraphics[width=0.32\linewidth]{prediction/simuNc4_adults_amplitudeSaccade_A1}}      
    
    \setcounter{subfigure}{0}

    \subfigure[2 year-old]{\includegraphics[width=0.32\linewidth]{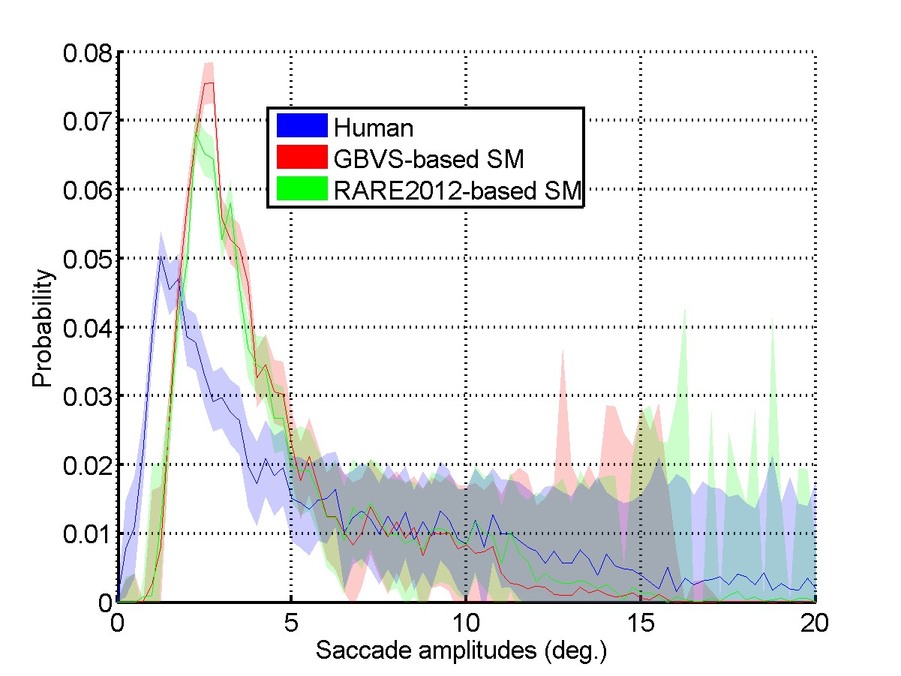}}        
    \subfigure[4-6 year-old]{\includegraphics[width=0.32\linewidth]{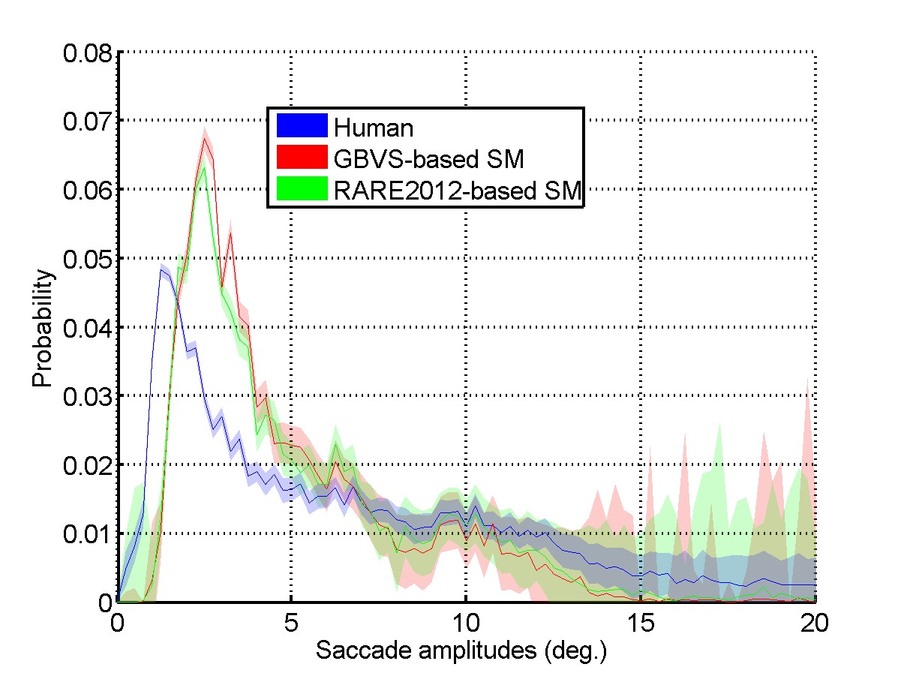}}    
    \subfigure[Adults]{\includegraphics[width=0.32\linewidth]{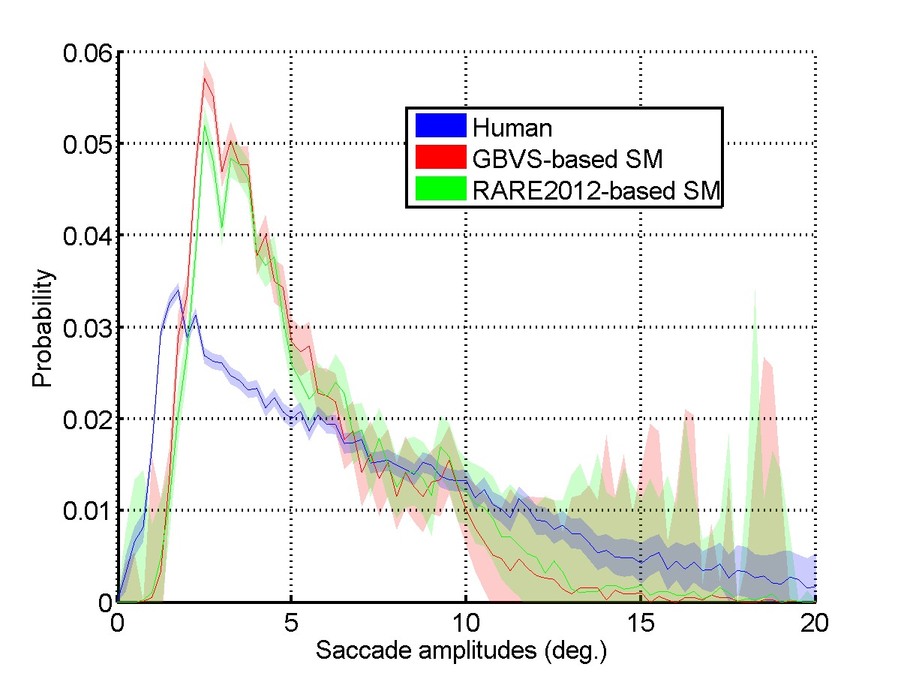}}    
    
  \captionof{figure}{Actual and predicted distributions of saccade amplitudes for 2 year-old (a), 4-6 year-old (b) and adults groups (c) for different values of $N_c$. Top row: $N_c=1$; middle row: best $N_c$; bottom row: $N_c=9$. Results are given for GBVS-based saccadic model and RARE2012-based saccadic model.}\label{fig:predictedScanpathsDistribution}
\end{figure*}

In this section, we demonstrate the influence of the parameter $N_c$ on the plausibility of the generated scanpaths. We remind that a low value of $N_c$ implies a high dispersion between observers due to an higher randomness. Increasing the value of $N_c$ reduces the randomness or the stochasticity of the model. In section 4.2.1 of the paper, we observe that the value of $N_c$ can be modified in the range from 3 to 7 without a significant loss of performance. It turns out that the parameter $N_c$ is important for generating plausible scanpaths. \figurename~\ref{fig:predictedScanpathsDistribution} illustrates this point. For 2 year-old, 4-6 year-old and adult groups, we plot the distributions of saccade amplitudes for different values of $N_c$: $N_c=1$ for the top row, $N_c=9$ for the bottom row and for the middle row, we choose the value of $N_c$ providing the best prediction: $N_c=5$ for 2 year-old and 4-6 year-old groups and $N_c=4$ for adult groups.

When $N_c=1$ (top row of \figurename~\ref{fig:predictedScanpathsDistribution}), the distribution of saccade amplitudes is almost flat. There is a high proportion of large saccades compared to the actual distribution of saccade amplitudes. When $N_c=9$ (bottom row of \figurename~\ref{fig:predictedScanpathsDistribution}), we observe an heavy-tailed distribution that could match with human distributions. However, the saccadic model makes significantly smaller saccades than human. 

We also observe in \figurename~\ref{fig:predictedScanpathsDistribution} that the distributions of saccade amplitudes estimated from the scanpaths stemming from GBVS-based saccadic model and RARE2012-based saccadic model are very similar. This suggests that the ability to produce plausible scanpaths does not depend on the input saliency map. The value of $N_c$ is in this case the key parameter.

\section{Conclusion}\label{sec:conclusion}
%****************************************************
In this paper, we show that saccadic models can be tailored for different age groups. Our saccadic model combines low-level salience, memory effects and viewing biases. Low-level salience is computed by state-of-the-art bottom-up saliency models. Memory effects represent the inhibition-of-return mechanism which performs the inhibition of an attended location in order to foster the scene exploration. The last component, i.e. viewing biases,  provides fundamental information about how observers explore a visual scene. We show that these viewing biases evolve with the maturation of the visual system. We were able to capture differences in gaze behaviour between age groups with joint distributions of saccade amplitudes and orientations. This representation, which is learned from actual eye tracking data, turns out to be fairly different for 2-year-olds, 4-6 year-olds, 6-10 year-olds and adult observers. By using this age-based visual signature, we showed that the proposed age-dependent saccadic model outperforms not only GBVS and RARE2012 saliency models but succeeds in generating scanpaths that match actual eye tracking data. 

Obviously, the present saccadic model cannot fully account for the complex nature of overt visual attention. Although that the joint distribution of saccade amplitudes and orientations has a number of merits, it would be required to incorporate other known properties of gaze behavior, such as the fixation duration, the time dependencies between successive saccades and advanced scanpath statistics. These aspects will be tackled in future works.

This study may have a significant impact on some computer vision applications. For instance, it would allow to tailor saliency-based image compression algorithms for observers of a specific age. Another example is related to image retargeting methods, which consist in reducing the image size while keeping the most visually important areas~\cite{Rubinstein2010comparative}. Most retargeting methods are based on importance maps that indicate the locations to preserve. Retargeting results could be improved by computing age-dependent importance map.

A side result of this work concerns the better understanding of the maturation of the visual system from childhood to adulthood, which could help the design of new applications to help visually impaired people (e.g. people suffering from ARMD (Age-Related Macular Degenerescence)). 

In supplementary material, a video sequence showing the maturation of eye movement behavior with respect to saccade amplitudes and orientations is provided. This video shows the influence of aging on saccade amplitudes and orientations, spanning from childhood to adulthood.

  \section*{Acknowledgment}

This work was supported in part by the National Natural Science Foundation of China under Grant No. 61471230.

\ifCLASSOPTIONcaptionsoff
  \newpage
\fi



\bibliographystyle{IEEEtran}
\bibliography{IEEEabrv,LeMeurSaccadicModel}
\end{document}